%% file: main.tex
\documentclass[10pt,twocolumn,letterpaper]{article}

\usepackage{iftex}
\ifPDFTeX\else\usepackage[T1]{fontenc}\fi

\usepackage[pagenumbers]{wacv}

\usepackage{amsmath}
\usepackage{amssymb}
\usepackage{booktabs}
\usepackage{longtable}
\usepackage{multirow}
\usepackage{graphicx}
\usepackage{xcolor}
\usepackage{colortbl}
\usepackage{array}
\usepackage{pifont}
\definecolor{oursrow}{rgb}{0.90,0.94,0.98}
\definecolor{bestcell}{gray}{0.88}

\definecolor{linkblue}{rgb}{0.21,0.49,0.74}
\usepackage[pagebackref,breaklinks,colorlinks,allcolors=linkblue]{hyperref}

\title{Per-Stroke Temporal Control for Text-to-Motion\\
via Action Units and Action-Detection Guidance}

\author{Euijun Jung \qquad Youngki Lee\\
Seoul National University\\
{\tt\small \{onsaga1, youngkilee\}@snu.ac.kr}
}

\begin{document}
\maketitle

\input{sec/0_abstract}
\input{sec/1_intro}
\input{sec/2_related}
\input{sec/3_method}
\input{sec/4_experiments}

\input{sec/5_discussion}
\input{sec/6_conclusion}

{
    \small
    \bibliographystyle{ieeenat_fullname}
    \bibliography{main}
}

\clearpage
\maketitlesupplementary
\appendix
\input{sec/x_appendix}

\end{document}

%% file: sec/0_abstract.tex
\begin{abstract}
Text-to-motion models are competent at the action a prompt names but unreliable at \emph{when} each stroke lands: four punches alternating left and right rarely return four separable strokes. We introduce typed temporal events called Action Units (AUs) that make the individual stroke --- its body track, action class, time window, and impact timing --- an explicit conditioning signal. We ground a frozen text-to-motion backbone on the AU set through a lightweight gated adapter injecting two streams (per-stroke tokens and a per-frame phase channel), and at inference close residual timing errors with a training-free classifier gradient from a frozen frame-level detector. We measure per-stroke control on StrokeBench, whose prompts specify count, ordering, track, and core-frame placement, paired with an audited stroke corpus. AU grounding markedly raises the rate of correctly placed single strokes over the strongest prior interface, at the best motion quality among text-, interval-, and frame-level baselines. The prompted core frame emerges as a further steerable axis.
\end{abstract}

%% file: sec/1_intro.tex
\section{Introduction}
\label{sec:intro}

Text-to-motion (T2M) generation is becoming a practical way to author 3D character animation from language, with applications spanning games, film, XR, and 3D content creation~\cite{zhu2024humanmotionsurvey}; diffusion- and token-based backbones trained on large captioned corpora~\cite{guo2022humanml3d,plappert2016kit} now render common actions convincingly~\cite{tevet2023mdm,chen2023mld,guo2024momask,zhang2024motiondiffuse,zhang2023t2mgpt,hong2025salad,zhang2023remodiffuse}. What they render reliably, however, is the action a caption \emph{names} --- not its temporal structure. Ask for ``four punches, alternating left and right,'' or ``a kick, then two quick punches, then a jump,'' and the named actions appear, but the user cannot control how many strokes occur, in what order, on which limb, or when each one lands.

\begin{figure}[t!]
\centering
\includegraphics[width=\linewidth]{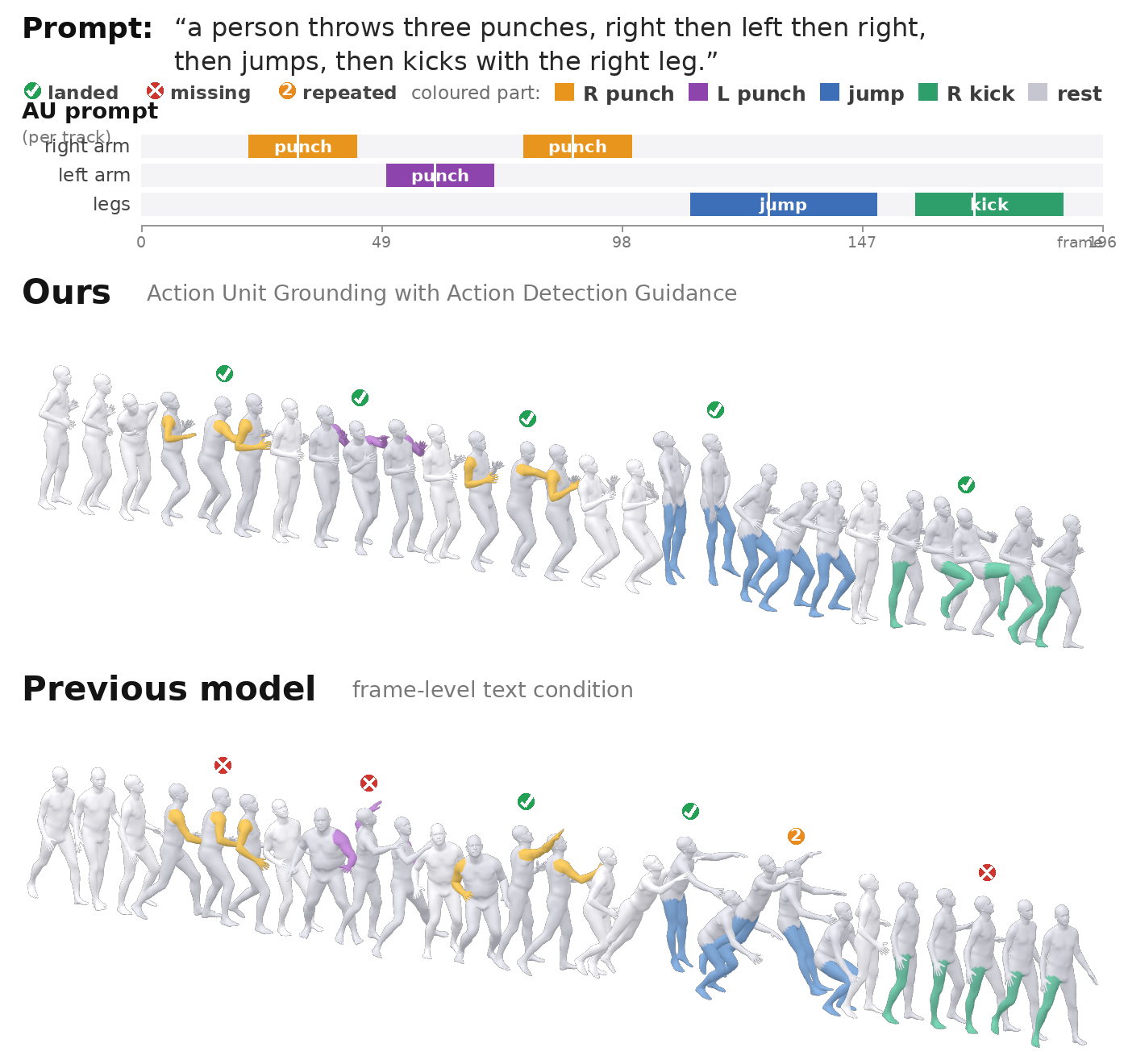}
\caption{\textbf{The Action Unit (AU) makes per-stroke structure an explicit condition.} The AU prompt (top) names each stroke's count, ordering, track, and core frame; conditioned on it, our frozen-backbone grounding (\emph{Ours}) places three alternating punches (R, L, R), a jump, then a kick. \textbf{Only each AU's responsible limb is colored} (punching arm L/R, kicking leg, both legs for the jump; rest gray); a marker above each head is the held-out detector's verdict --- \textbf{green} in-window on track, \textbf{red} missing, \textbf{amber ``2''} fired twice. Ours lands every stroke on the right limb in-window (the jump aided by our loft prior, \S\ref{sec:method}); a frame-level baseline~\cite{li2025unimotion} on the same text lands one of three punches and \emph{jumps twice}. SMPLify-3D meshes.}
\label{fig:teaser}
\end{figure}

The established way to request such control is \emph{interval-level} conditioning: the user authors a timeline of per-interval sub-captions and a model composes them~\cite{athanasiou2022teach,zhang2023finemogen,petrovich2024stmc,li2026frankenmotion,li2025unimotion,zhao2025dartcontrol}. But the unit of control is the \emph{interval} (or body part), not the individual stroke, and two gaps remain: reaching one stroke means authoring a sub-caption per stroke and stitching many short clips, whose actions spread to fill --- and bleed past --- their edges so neighbours run together; and stitching separately denoised intervals leaves seams that read as unnatural transitions~\cite{athanasiou2022teach,petrovich2024stmc}, buying placement at a cost in whole-sequence coherence. What is missing is a control primitive at the level of the individual stroke.

We make the individual stroke the addressable primitive: the Action Unit (AU), a typed temporal event that carries one stroke's body track, action class, time window, and its load-bearing \emph{core frame} --- the instant the action lands (impact for a punch or kick, apex for a jump, release for a throw, reception for a catch). A motion is the ordered set of its AUs, so count, ordering, and concurrency are read directly off the schema. The core frame addresses what even a cleanly placed interval leaves out: the moment \emph{within} the stroke a prompt may care about, which no prior interface exposes. Our focus is to make a frozen generator honor this condition --- place each stroke in its prompted window, with the right count and side, and keep it out of the gaps --- and, where the corpus leaves room, to steer the core frame within that window.

We obtain this control without retraining the generator. A lightweight gated adapter grounds a frozen text-to-motion backbone~\cite{tevet2023mdm} on the AU set --- per-stroke tokens plus a per-frame within-stroke phase channel --- and already seats most strokes correctly at training time. Action-Detection Guidance (ADG) then finishes placement at inference. It reads a frozen frame-level detector on the running clean-motion estimate, and in one classifier gradient fills under-filled windows and silences leaked classes. Neither stage retrains the generator or regresses motion quality.

We evaluate on StrokeBench, our per-stroke benchmark over eight action classes. The contributions are:
(i) the Action Unit --- a typed per-stroke temporal event carrying a within-stroke core frame --- together with the AU corpus of audited strokes, which we will release;
(ii) Action Unit Grounding, a lightweight, portable gated adapter --- per-stroke tokens plus a per-frame phase channel --- that grounds a frozen backbone on the AU set, realizes the prompted count, side, and stroke placement, and ports unchanged across frozen transformer backbones;
(iii) Action-Detection Guidance, a training-free classifier gradient from a frozen frame-level detector that fills under-filled windows and silences off-AU leakage at inference;
and (iv) StrokeBench, a per-stroke benchmark with per-AU correctness metrics.

%% file: sec/2_related.tex
\section{Related Work}
\label{sec:related}

Controlling \emph{when} actions occur in a generated motion has been pursued through ever finer control units, yet none reaches the individual stroke: even a per-frame label tags \emph{what} happens at a frame without making the stroke --- its count, side, and within-stroke timing --- an addressable primitive. We review prior interfaces by their control unit; Table~\ref{tab:granularity} situates them by that unit and within-stroke reach.

\begin{table*}[tb]
\centering
\footnotesize
\caption{Conditioning interfaces for text-to-motion control, by broad category. \emph{Interface}: the form of the conditioning signal the user supplies. \emph{Control unit}: the temporal granularity at which it places content. Only the Action Unit reaches the individual stroke; addressing the within-stroke core frame is unique to it.}
\label{tab:granularity}
\setlength{\tabcolsep}{6pt}
\begin{tabular}{@{}p{2.6cm} p{9.6cm} c@{}}
\toprule
Interface & Representative methods & Control unit \\
\midrule
Global caption & MDM~\cite{tevet2023mdm}, MLD~\cite{chen2023mld}, MoMask~\cite{guo2024momask}, T2M-GPT~\cite{zhang2023t2mgpt}, MotionDiffuse~\cite{zhang2024motiondiffuse}, SALAD~\cite{hong2025salad}, ReMoDiffuse~\cite{zhang2023remodiffuse} & whole motion \\
Joint trajectory & OmniControl~\cite{xie2024omnicontrol}, CondMDI~\cite{cohan2024condmdi}, GMD~\cite{karunratanakul2023gmd}, Kimodo$^*$~\cite{kimodo2026} & per-frame \\
Interval-/frame-level text & TEACH~\cite{athanasiou2022teach}, STMC~\cite{petrovich2024stmc}, FineMoGen~\cite{zhang2023finemogen}, FrankenMotion~\cite{li2026frankenmotion}, UniMotion~\cite{li2025unimotion}, DART~\cite{zhao2025dartcontrol}, FineXtrol~\cite{shen2026finextrol} & interval/frame \\
\midrule
\textbf{Action Unit (ours)} & one typed temporal event per stroke & per-stroke event \\
\bottomrule
\end{tabular}
\\[3pt]
{\footnotesize $^*$Kimodo also exposes an interval-level text timeline.}
\end{table*}

\paragraph{Text-to-motion with global caption.}
Text-to-motion backbones map a single global caption to motion --- diffusion~\cite{tevet2023mdm,chen2023mld,zhang2024motiondiffuse,hong2025salad,zhang2023remodiffuse,dabral2023mofusion,kim2023flame,yuan2023physdiff} and discrete-token autoregressive models~\cite{guo2024momask,zhang2023t2mgpt,jiang2023motiongpt,zhong2023attt2m,pinyoanuntapong2024mmm,pinyoanuntapong2024bamm} now render common actions convincingly, and some enrich \emph{what} a caption specifies without changing its granularity (e.g.\ hierarchical action/sub-action parsing~\cite{jin2023actaswish}). The unit of control stays the whole sentence: text selects which actions appear and their coarse style, but the individual stroke is not addressable.

\paragraph{Interval/frame-level text control.}
A complementary line lets the user author a timeline of sub-captions. TEACH~\cite{athanasiou2022teach} composes \emph{sequential} sub-action texts, and STMC~\cite{petrovich2024stmc} extends per-interval action text to parallel per-body-part timelines. FineMoGen~\cite{zhang2023finemogen} routes per-region timing language through a spatio-temporal mixture. FrankenMotion~\cite{li2026frankenmotion} composes body-part atoms with explicit LLM-derived time windows. UniMotion~\cite{li2025unimotion} drops to per-frame labels in a unified motion--text diffusion model, and DART~\cite{zhao2025dartcontrol} sequences short text-labeled primitives autoregressively. FineXtrol~\cite{shen2026finextrol} embeds time markers in the caption. These place text at the interval or per-frame label, but none exposes the individual stroke as an addressable primitive with its own track, laterality, and within-stroke timing, from which count and ordering follow. Pushing them toward per-stroke control means authoring ever-shorter intervals, and composition degrades. We compare against each through its native interface (\S\ref{sec:main}).

\paragraph{Injecting a condition into a frozen backbone.}
Two recipes add a new \emph{trained} module to a frozen generator without touching its weights. The \emph{control-branch} recipe of ControlNet~\cite{zhang2023controlnet} clones the trunk into a trainable copy; in T2M, OmniControl~\cite{xie2024omnicontrol} pairs such a branch (its realism module) with sparse spatial joint targets placed at chosen frames. The \emph{gated-injection} recipe instead adds zero-gated attention over a few condition tokens to the frozen blocks, as in GLIGEN~\cite{li2023gligen} for box-grounded images and gated key/value memories for motion joint trajectories~\cite{sun2026kvcontrol}. Both carry spatial or coarse whole-action signals, not a per-stroke event. As our condition is a handful of typed temporal events, we inject the per-stroke tokens by gated attention and add a per-frame phase residual for the within-stroke channel (\S\ref{sec:grounding}), benchmarking a control-branch variant under an identical conditioning signal and protocol (\S\ref{sec:roles}).

\paragraph{Classifier guidance at sampling time.}
A classifier gradient steers a frozen diffusion model toward an attribute without retraining~\cite{dhariwal2021diffusionbeatsgan}; training-free variants read the guide on the predicted clean motion $\hat{\mathbf{x}}_0$~\cite{chung2023dps,yu2023freedom,song2023pigdm} but are unstable at high noise. In T2M this route imposes spatial constraints without a trained branch --- imputation and analytic guidance for keyframes and trajectories~\cite{cohan2024condmdi,karunratanakul2023gmd}. We instead use a frozen frame-level action detector as the classifier, with per-sample gradient normalization for stability, letting one gradient both fill under-filled windows and silence off-AU leakage (\S\ref{sec:adg}).

%% file: sec/3_method.tex
\section{Method}
\label{sec:method}

\begin{figure*}[t]
\centering
\includegraphics[width=\textwidth]{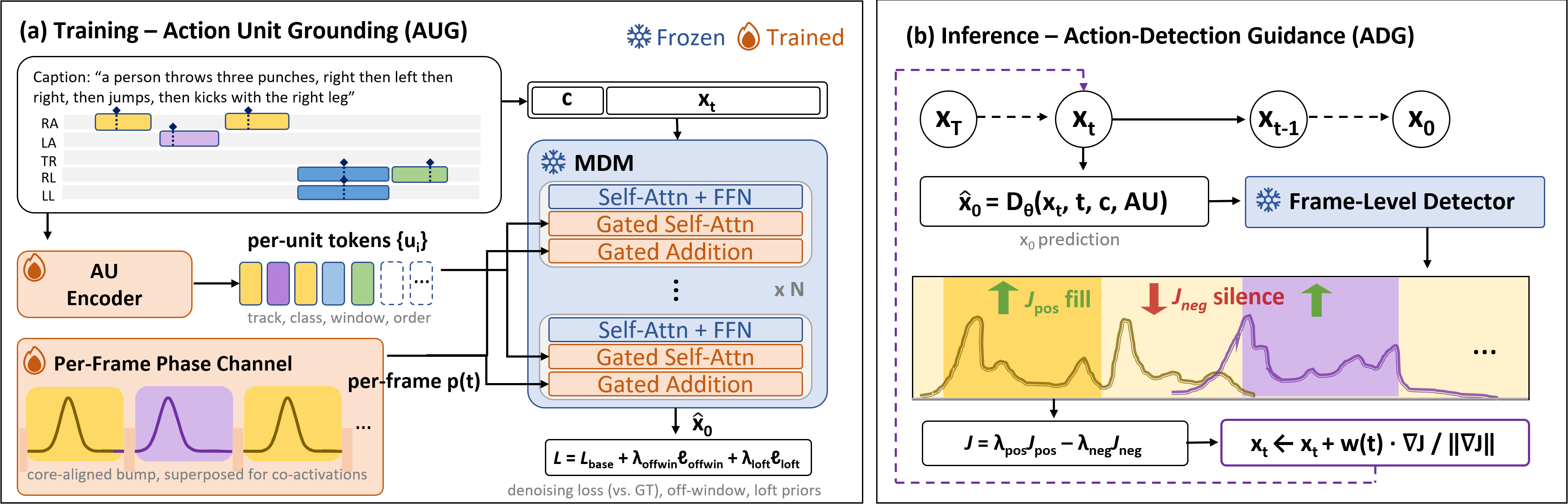}
\caption{Pipeline. \textbf{(a) Training:} a gated self-attention adapter injects two streams into the \emph{frozen} MDM --- a per-stroke AU token set (each with its window) and a per-frame within-stroke phase channel (Eq.~\ref{eq:phase-field}) --- under a fixed off-window velocity prior suppressing responsible-joint motion in the gaps. Both gates are zero-initialized, so with no AU the adapter collapses \emph{exactly} to the backbone. \textbf{(b) Inference:} a \emph{separate} frozen frame-level detector reads the predicted clean motion and, via Action-Detection Guidance (ADG), applies one classifier gradient that fills under-filled windows and silences leaked off-AU classes --- no weights trained.}
\label{fig:pipeline}
\end{figure*}

\subsection{Overview}
\label{sec:overview}
Our pipeline (Fig.~\ref{fig:pipeline}) makes per-stroke structure controllable around a single \emph{frozen} text-to-motion backbone~\cite{tevet2023mdm}. The \textbf{Action Unit} (AU) is the representation --- one typed temporal event that makes the individual stroke, core frame included, addressable (corpus built by human audit, \S\ref{sec:corpus}). \textbf{Action Unit Grounding} (AUG, \S\ref{sec:grounding}) is the trained interface: a lightweight gated adapter injecting the AU set into the frozen denoiser as two streams --- per-stroke tokens and a per-frame phase channel --- under a fixed off-window prior. \textbf{Action-Detection Guidance} (ADG, \S\ref{sec:adg}) corrects at inference: a \emph{separate} frozen frame-level detector (\S\ref{sec:detector}) that, in one classifier gradient and no further training, fills under-filled windows and silences leaked gaps, and whose per-frame class signal StrokeBench also reads.

\subsection{Action Unit Corpus}
\label{sec:corpus}
We build the AU corpus by human audit, drawing motions from HumanML3D~\cite{guo2022humanml3d} and, for the held-out partition, FrankenMotion~\cite{li2026frankenmotion} (AMASS body-part atoms). An action segmenter trained on BABEL frame labels~\cite{punnakkal2021babel,farha2019mstcn} surfaces candidate windows, velocity peaks on the responsible joint split each candidate into per-stroke proposals, and an annotator accepts, boundary-edits, or rejects each, tagging every kept stroke with its track, class, window, and core frame; mirror augmentation doubles left/right coverage. The eight-class corpus --- six base classes (punch, kick, throw, jump, hop, catch) plus one directional pair, squat (down/up) --- splits by HumanML3D partition into \textbf{AUC-T} (the \emph{train} partition, ${\sim}4{,}400$ Action Units over ${\sim}1{,}039$ clips), which trains our model and the ADG guide, and a disjoint held-out partition \textbf{AUC-E} (HML test/val plus FrankenMotion, ${\sim}3{,}200$ AUs over $711$ clips) used only to train the StrokeBench evaluator (\S\ref{sec:setup}, Supp.~\ref{app:heldout}); see Table~\ref{tab:corpus}. Proposal filtering and the audit interface are detailed in Supp.~\ref{app:corpus}.

\begin{table}[t]
\centering\footnotesize
\setlength{\tabcolsep}{5pt}
\begin{tabular}{l rr rr r}
\toprule
 & \multicolumn{2}{c}{AUC-T} & \multicolumn{2}{c}{AUC-E} & \\
\cmidrule(lr){2-3}\cmidrule(lr){4-5}
Class & AUs & clips & AUs & clips & Dur\\
\midrule
punch & 1{,}062 & 216 & 478 &  47 & 15\\
kick  &   964 & 280 & 656 & 192 & 23\\
throw &   628 & 243 & 604 & 199 & 22\\
jump  &   592 & 120 & 452 & 175 & 13\\
hop   &   324 &  22 & 104 &   9 & 10\\
catch &   230 &  63 & 550 &  68 & 15\\
squat\_down & 346 & 127 & 216 & 78 & 23\\
squat\_up   & 260 &  88 & 176 & 60 & 21\\
\midrule
Total & 4{,}406 & 1{,}039 & 3{,}236 & 711 & ---\\
\bottomrule
\end{tabular}
\caption{The eight-class AU corpus, per class: Action Units (after mirror augmentation) and distinct source clips, by HumanML3D partition. \textbf{AUC-T} (HML-\emph{train}) trains the adapter $+$ ADG guide; the disjoint held-out \textbf{AUC-E} (HML test/val $+$ FrankenMotion, $\notin$ training) trains only the evaluator (Supp.~\ref{app:heldout}). A clip may serve several classes, so per-class clips exceed the distinct totals. \emph{Dur}: median window duration (frames, 20\,fps).}
\label{tab:corpus}
\end{table}

\subsection{Action Unit Grounding}
\label{sec:grounding}

\paragraph{Conditioning inputs.}
The AU set enters as \emph{two streams} --- a per-stroke \emph{token} carrying the stroke's discrete layout, and a per-frame \emph{phase channel} carrying its within-window execution --- which we describe in turn.

\emph{(i) Per-stroke token.} \emph{Each} of the $K$ prompted AUs $e_i=(g_i,s_i,t_s^i,t_c^i,t_e^i)$ --- body track, action class, and start/core/end frames --- is encoded by the \emph{same} encoder $f_{\mathrm{AU}}$ into its own $D$-dimensional token,
\begin{equation}
\mathbf{u}_i = f_{\mathrm{AU}}\!\left(\bigl[\phi(t_s^i)\,\|\,\phi(t_c^i)\,\|\,\phi(t_e^i)\,\|\,\mathbf{w}_{g_i}\,\|\,\mathbf{a}_{s_i}\,\|\,\nu_i\,\|\,f_{\mathbf{c}}(\mathbf{c})\bigr]\right),
\label{eq:au-token}
\end{equation}
applied to each of the $K$ prompted strokes, so they form a \emph{token set} $\{\mathbf{u}_i\}_{i=1}^{K}$ (Eq.~\ref{eq:au-token} is the per-stroke encoder; the whole set is attended \emph{jointly} in Eq.~\ref{eq:gate}, not one token at a time), with Fourier-expanded frame fractions $\phi$, learned track/class embeddings $\mathbf{w}_g,\mathbf{a}_s$, a within-track instance index $\nu_i$, and a linear read-out $f_{\mathbf{c}}$ of the global text embedding $\mathbf{c}$, fused by a two-layer MLP $f_{\mathrm{AU}}$. This token is mostly \emph{structural} --- timing, track, count, and a class label; the per-stroke class embedding $\mathbf{a}_s$ localizes class to a window, the global-caption read-out $f_{\mathbf{c}}$ is a linear projection that re-reads the same global text embedding $\mathbf{c}$ the backbone already consumes (so each stroke is caption-aware), and the time fields use a frequency code (each field is ablated in Supp.~\ref{app:knockout}). Per-slot existence flags let the active-slot count equal the supplied stroke count, so count and ordering are read directly off the set (slot budget in Supp.~\ref{app:hparams}).

\emph{(ii) Per-frame phase channel.} $\mathbf{p}(t)$ marks \emph{where within each window} the load falls. Each stroke active at frame $t$ contributes a learned per-frame vector $\mathbf{p}_i(t)$ --- a duration-aware Gaussian bump centered on its core $t_c^i$, fused with build/follow-through state flags and the track/class embeddings (Supp.~\ref{app:hparams}) --- and the channel is their count-normalized \emph{superposition} over the strokes active at that frame,
\begin{equation}
\mathbf{p}(t)=\frac{1}{n(t)}\sum_{i\,\in\,\mathcal{A}(t)}\mathbf{p}_i(t),\qquad \mathcal{A}(t)=\{\,i:\,t\in[t_s^i,t_e^i]\,\},
\label{eq:phase-field}
\end{equation}
where $\mathcal{A}(t)$ is the set of strokes whose window contains frame $t$ and $n(t){=}|\mathcal{A}(t)|$ their count, so co-active strokes' contents \emph{superpose} (count-normalized) rather than overwrite. This superposition is the mechanism behind concurrent placement: two strokes overlapping in time on different tracks each deposit their own phase content into the same frames, so neither is lost --- which the ablation confirms is the dominant carrier of concurrency (\S\ref{sec:overlap}), more so than the discrete tokens. The token set supplies the layout --- \emph{which} strokes exist and where --- and the superposed phase channel their continuous, co-active within-window execution; together they place each stroke.

\paragraph{Conditioning injection.}
Writing $D_\theta$ for the frozen $\mathbf{x}_0$-predicting denoiser, the adapter realizes the AU-conditioned model $\hat{\mathbf{x}}_0 = D_\theta(\mathbf{x}_t,\,t,\,\mathbf{c},\,\{\mathbf{u}_i\},\,\mathbf{p})$ by \emph{additively} injecting the two AU streams into every frozen transformer block of the backbone~\cite{tevet2023mdm}. At block $l$ and motion frame $t$, the block feature $\mathbf{h}^{(l)}_t$ first takes a gated self-attention over the per-stroke tokens~\cite{li2023gligen,wang2024instancediffusion}, read off the block feature \emph{before} the phase residual,
\begin{equation}
\mathbf{h}^{(l)}_t \leftarrow \mathbf{h}^{(l)}_t + \tanh\!\big(\gamma^{(l)}_{\mathbf{u}}\big)\,\big[\mathrm{Attn}\!\big(\mathbf{h}^{(l)},\,\{\mathbf{u}_i\}\big)\big]_t,
\label{eq:gate}
\end{equation}
then a gated per-frame residual from the phase channel,
\begin{equation}
\mathbf{h}^{(l)}_t \leftarrow \mathbf{h}^{(l)}_t + \tanh\!\big(\gamma^{(l)}_{\mathbf{p}}\big)\,\mathbf{p}_t,
\label{eq:gate-p}
\end{equation}
where $\gamma^{(l)}_{\mathbf{u}},\gamma^{(l)}_{\mathbf{p}}$ are per-block learned gates initialized at $0$, so each stream starts inert and the frozen prior is recovered exactly. The token attention is unmasked: each token carries its own window through the encoded boundaries $\phi(t_s^i),\phi(t_c^i),\phi(t_e^i)$, and binding a stroke to its window --- with within-window placement --- is resolved by the per-frame phase channel (Eq.~\ref{eq:phase-field}, which sums only the strokes active at $t$) together with the off-window prior, not by a hard attention mask (Supp.~\ref{app:knockout} explores adding an explicit window mask or Gaussian bias). Only the adapter and the AU encoder are trained; the AU stream is dropped with probability $p{=}0.3$ for classifier-free guidance.

\paragraph{Training objective.}
We train the adapter and AU encoder with the backbone's denoising loss $\mathcal{L}_{\mathrm{base}}$ --- its $\mathbf{x}_0$-prediction objective, now conditioned on the AU stream --- plus a fixed off-window prior and a class-conditional loft prior,
\begin{equation}
\mathcal{L}=\mathcal{L}_{\mathrm{base}}+\lambda_{\mathrm{offwin}}\,\ell_{\mathrm{offwin}}+\lambda_{\mathrm{loft}}\,\ell_{\mathrm{loft}}.
\label{eq:objective}
\end{equation}
$\ell_{\mathrm{offwin}}$ ($\lambda_{\mathrm{offwin}}{=}0.8$) suppresses responsible-joint velocity in every non-AU gap, and $\ell_{\mathrm{loft}}$ ($\lambda_{\mathrm{loft}}{=}5$) lifts the one cue the velocity proxies underconstrain --- airborne height --- by requiring each jump/hop AU's core-frame pelvis height to clear the clip's off-window mean by a hinged margin $m{=}0.10$\,m (Supp.~\ref{app:objective}). Both priors are active in the headline configuration; $\ell_{\mathrm{loft}}$ affects only jump and hop.

\subsection{Action-Detection Guidance}
\label{sec:adg}
The trained conditioning already seats the prompted stroke in its window, but two residuals remain: a prompted window with no detectable stroke of its class (\emph{under-recall}), and a prompted class surfacing in a gap (\emph{off-AU leakage}). We close both at inference, touching no trained weight, by reading a frozen frame-level detector (\S\ref{sec:detector}) on the sampler's current clean-motion estimate and ascending a weighted sum of two per-AU terms,
\begin{equation}
\mathcal{J} = \lambda_{\mathrm{pos}}\,\mathcal{J}_{\mathrm{pos}} \;-\; \lambda_{\mathrm{neg}}\,\mathcal{J}_{\mathrm{neg}},
\label{eq:adg-obj}
\end{equation}
an in-AU recall term $\mathcal{J}_{\mathrm{pos}}$ that saturates once a window reads as filled, and an off-AU term $\mathcal{J}_{\mathrm{neg}}$ that silences the class on gap runs. Read on the clean-motion estimate, this gradient is numerically unstable at high noise; we apply a per-sample gradient-norm normalization (following DPS~\cite{chung2023dps}) so the update stays bounded at every noise level (terms and stabilization in Supp.~\ref{app:adg}).

\subsection{Unit Detector}
\label{sec:detector}
We use \emph{two different} frozen detectors --- one to steer ADG in-loop, a separate one to score StrokeBench --- so the reported metric does not reuse the network that steered generation. Both are frame-level temporal action segmenters (per-frame multi-label, not clip-level recognition): they map the $T{\times}263$ HumanML3D feature to per-frame, per-class presence through a per-class sigmoid, over a class set that splits each laterally-typed action into left/right (footedness legible) and each directional pair into down/up. The ADG guide is a convolutional \textbf{C2F-TCN}~\cite{singhania2023c2ftcn} trained on \textbf{AUC-T}, whose locally smooth gradients suit per-step guidance; the StrokeBench evaluator is a dilated-attention \textbf{ASFormer}~\cite{yi2021asformer} trained on the held-out \textbf{AUC-E} --- a different architectural family, so guidance does not inflate the metric. Training recipe, feature layout, and the cross-source mapping for out-of-HumanML3D motions are in Supp.~\ref{app:detector}.

%% file: sec/4_experiments.tex
\section{Experiments}
\label{sec:exp}

\begin{table*}[t]
\centering
\scriptsize
\setlength{\tabcolsep}{4pt}
\caption{StrokeBench \textbf{eight-class} comparison: \textbf{prompt-weighted} mean over the seven single-class sub-benches ($10$ samples/prompt; Ours averaged over $3$ training seeds). Best per column \textbf{bold}, F1\textsubscript{AU} winner shaded; last two columns are F1\textsubscript{AU} on the compositional chain/overlap sub-benches. Metrics \S\ref{sec:strokebench}; FID/DIV reference and baseline conversion \S\ref{sec:setup}, Supp.~\ref{app:stmc}.}
\label{tab:main}
\begin{tabular}{llcccccc>{\hspace{4pt}}c>{\hspace{6pt}}cc}
\toprule
 & & \multicolumn{7}{c}{Single-class (eight-class mean)} & \multicolumn{2}{c}{Comp.\ F1\textsubscript{AU}$\uparrow$}\\
\cmidrule(lr){3-9}\cmidrule(l){10-11}
Interface & Method & F1\textsubscript{AU}$\uparrow$ & IUE$\uparrow$ & IUR$\uparrow$ & GLR$\downarrow$ & FID$\downarrow$ & DIV$\,\to$ & F1@.25$\uparrow$ & chain & overlap\\
\midrule
Global caption & MDM~\cite{tevet2023mdm} & 0.544 & 0.423 & 0.478 & 0.236 & 27.91 & 2.79 & 0.269 & 0.508 & 0.508\\
\midrule
\multirow{2}{*}{\shortstack[l]{Joint\\trajectory}}
 & OmniControl~\cite{xie2024omnicontrol} & 0.637 & 0.512 & 0.590 & 0.157 & 25.61 & 2.73 & 0.475 & 0.616 & 0.590\\
 & Kimodo$^{\dagger\ddagger}$~\cite{kimodo2026} & 0.839 & 0.757 & 0.758 & 0.058 & 35.24 & 1.84 & 0.754 & 0.700 & 0.687\\
\midrule
\multirow{6}{*}{\shortstack[l]{Interval-/frame-\\level text}}
 & FineMoGen~\cite{zhang2023finemogen} & 0.626 & 0.460 & 0.479 & \textbf{0.020} & 33.62 & 1.32 & 0.610 & 0.490 & 0.463\\
 & STMC~\cite{petrovich2024stmc}, standstill & 0.757 & 0.644 & 0.644 & 0.081 & 32.10 & 1.64 & 0.670 & 0.755 & 0.442\\
 & STMC, overlay & 0.646 & 0.484 & 0.485 & 0.028 & \textbf{21.52} & 3.31 & 0.597 & 0.608 & 0.587\\
 & UniMotion~\cite{li2025unimotion} & 0.800 & 0.693 & 0.703 & 0.054 & 25.25 & 2.35 & 0.707 & 0.800 & 0.699\\
 & DART~\cite{zhao2025dartcontrol} & 0.672 & 0.537 & 0.592 & 0.101 & 25.17 & 4.25 & 0.548 & 0.480 & 0.305\\
 & FrankenMotion$^{\dagger}$~\cite{li2026frankenmotion} & 0.602 & 0.471 & 0.538 & 0.165 & 27.77 & 3.38 & 0.400 & 0.539 & 0.418\\
\midrule
Unit grounding (ours) & AUG $+$ ADG & \cellcolor{oursrow}\textbf{0.898} & \textbf{0.886} & \textbf{0.894} & 0.087 & 27.80 & 2.07 & \textbf{0.808} & \textbf{0.829} & \textbf{0.767}\\
\bottomrule
\end{tabular}
\\[3pt]{\footnotesize $^\dagger$Cross-dataset (not HumanML3D): FrankenMotion (AMASS body-part atoms, $1$ deterministic motion/prompt), Kimodo (hosted checkpoint). $^\ddagger$Kimodo uses both native interfaces together (interval text timeline $+$ keypoint constraints). Ours $3$-seed std $\leq.04$ across columns.}
\end{table*}

\subsection{Setup}
\label{sec:setup}
On the frozen backbone we train only the conditioning adapter and AU encoder on AUC-T (\S\ref{sec:corpus}; 300 epochs, batch 8, AdamW $10^{-4}$; ${\sim}1.5$--$1.8$\,h on one NVIDIA A6000). At inference we apply classifier-free guidance (scale $2.5$) on the AU stream and ADG ($\lambda_{\mathrm{pos}}{=}2.0$, $\lambda_{\mathrm{neg}}{=}1.5$; Supp.~\ref{app:strength}). We compare against caption-only, interval-/frame-level text, and joint-trajectory baselines~\cite{tevet2023mdm,zhang2023finemogen,li2025unimotion,zhao2025dartcontrol,petrovich2024stmc,li2026frankenmotion,xie2024omnicontrol,kimodo2026}, each through its native interface (conversion in Supp.~\ref{app:stmc}). \textbf{FID and DIV} follow the HumanML3D protocol~\cite{guo2022humanml3d}: FID against a class-matched held-out reference --- all $\notin$-training clips of each class (HML test/val $+$ FrankenMotion) --- averaged over the seven single-class sub-benches; DIV within the generated set. Control metrics (\S\ref{sec:strokebench}) are pooled over all prompts of the seven single-class sub-benches --- a \emph{prompt-weighted} mean, summing the raw per-window counts across prompts before forming each ratio, so larger sub-benches weigh proportionally --- and read off the held-out AUC-E evaluator at $\tau{=}0.5$ (\S\ref{sec:detector}, Supp.~\ref{app:heldout}).

\subsection{StrokeBench}
\label{sec:strokebench}

StrokeBench scores per-stroke control on motions from AU-typed prompts in three regimes. \textbf{Single-class}: seven sub-benches, one per action class (squat scoring both down/up phases), each prompting one \emph{class} but possibly several strokes of it, isolating timing, count, and side within a class. \textbf{Chain}: cross-class \emph{sequential} prompts whose strokes do not overlap in time. \textbf{Overlap}: \emph{concurrent} prompts placing two strokes on different tracks whose windows \emph{partially} overlap in time (e.g.\ a punch begun mid-kick), not two strokes sharing one whole window. Each prompt specifies an AU set (count, ordering, track, core-frame placement) and a global caption; we sample $10$ motions per prompt and score per-stroke placement (Supp.~\ref{app:strokebench}), all read off the held-out detector (\S\ref{sec:detector}). A \emph{chunk} is a maximal contiguous run of frames the detector reads as one class; a class's \emph{Non-AU gap} is the complement of the union of its AU windows:
\begin{itemize}\setlength{\itemsep}{1.5pt}\setlength{\parskip}{0pt}
\item \textbf{Unit axes.} \textbf{IUR}/\textbf{IUE}: the fraction of windows holding $\geq1$ / \emph{exactly one} target-class chunk that overlaps the window. \textbf{GLR}: the fraction of Non-AU gaps holding a leaked chunk --- an \emph{extra} same-class stroke in the gap belonging to no prompted window, so a real stroke bleeding a frame past its boundary is not charged. \textbf{F1\textsubscript{AU}}: the harmonic mean of IUE and $1{-}$GLR, high only when strokes are both in-window and absent off-window. \textbf{F1@.25}: the segmental F1 matching detected to prompted segments at temporal IoU ${\ge}0.25$~\cite{lea2017tcn}.
\end{itemize}
The prompted core frame $t_c$ is a further axis, scored by the \textbf{proxy--core slope} $\hat s$. The realized core is read from a \emph{proxy} --- the responsible-joint velocity peak (airborne-ankle apex for jumps), which precedes the true core by a class-typical offset $\delta_g$, so the expected proxy frame is $t_c{+}\delta_g$. Regressing the detected proxy offset-free on the prompted $t_c$ fraction gives $\hat s$, with $\hat s{\to}1$ one-for-one tracking and $\hat s{\approx}0$ no control. It proves an emergent, steerable axis on five of the six non-postural classes (\S\ref{sec:coreframe}).

\subsection{Comparison to Other Methods}
\label{sec:main}

\begin{figure*}[t]
\centering
\includegraphics[width=\textwidth]{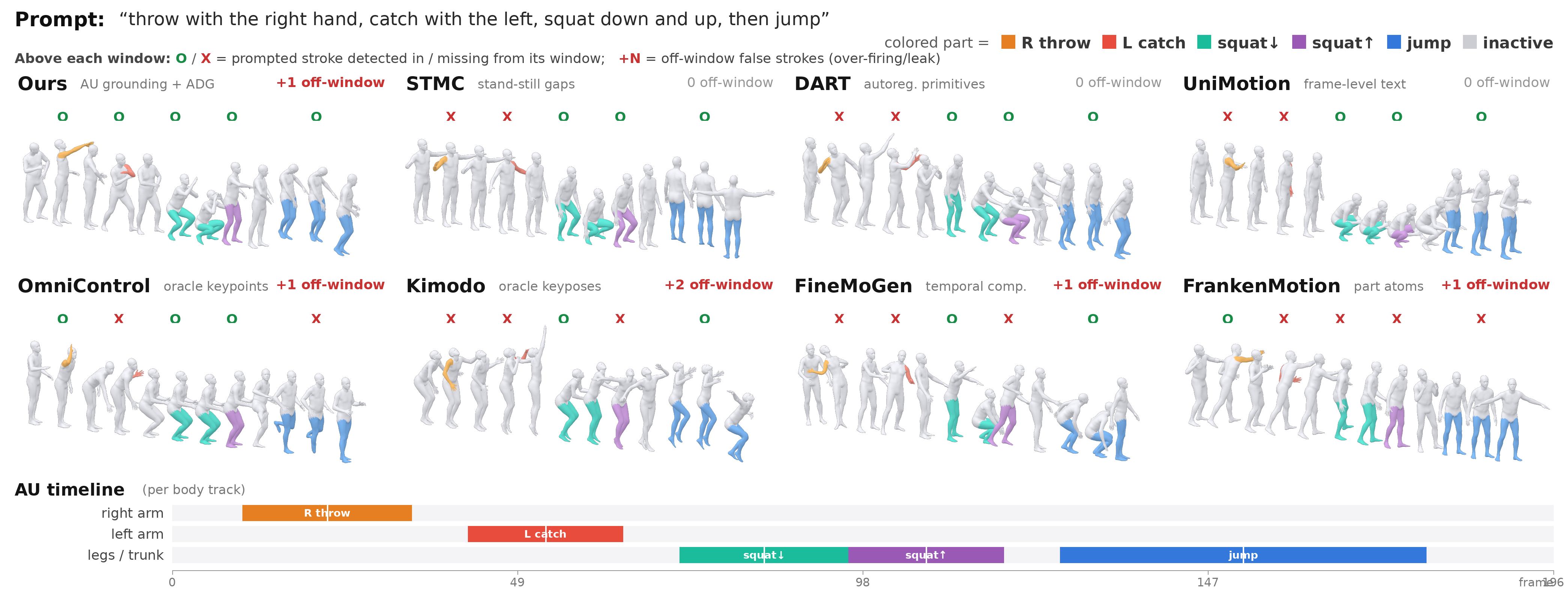}
\caption{A four-action chain --- \emph{``throw right, catch left, squat down/up, then jump''} --- by eight interfaces (SMPL meshes, temporal order). \textbf{Only each AU's responsible part is tinted} (throwing/catching arm; legs and trunk for squat and jump), so a colored limb inside its window means the \emph{correct} part moved at the \emph{prompted} time; above each window O/X is the held-out detector's verdict and red \textbf{$+N$} counts off-window false strokes. Ours is the only interface to land all five windows; the frame-/interval-level baselines (UniMotion, STMC, DART) recover the large squat and jump but drop the two subtle arm strokes, OmniControl places only the throw, Kimodo/FineMoGen part, FrankenMotion one. Short arm actions wedged between large lower-body motions are where competing interfaces fail.}
\label{fig:qual}
\end{figure*}

Table~\ref{tab:main} reports the eight-class single-class mean and the compositional chain/overlap sub-benches. Unit grounding leads every placement axis (F1\textsubscript{AU} $0.898$, IUE $0.886$, F1@.25 $0.808$); FID is a control--realism trade-off, ours ($27.8$, held-out AUC-E reference) sitting mid-pack between the low-motion interfaces that place few strokes and the over-constrained ones. The strongest HumanML3D-trained baseline, UniMotion ($0.80$), reaches its score only by ingesting our dense per-frame annotation and saturating on the slow squat phases; we beat it on every placement axis (IUE $0.886$ vs.\ $0.69$, F1@.25 $0.808$ vs.\ $0.71$). The cross-dataset Kimodo, through its \emph{full} keypoint$+$text interface, places strokes nearly as well ($0.84$) but only by overconstraining its generations off-distribution (FID $35.2$, a loose upper bound; Supp.~\ref{app:stmc}). The lead holds in both compositional regimes --- chain $0.83$ and overlap $0.77$, against UniMotion's $0.80$ and $0.70$ --- where per-frame text handles \emph{sequential} chains but cannot place \emph{simultaneous} strokes (\S\ref{sec:overlap}; read \emph{within} a column, as these sub-benches are punch/kick-heavy). It concedes GLR ($0.087$) only to methods that emit few detectable strokes (FineMoGen $0.020$, STMC-overlay $0.028$). Figure~\ref{fig:qual} renders one four-action chain where these differences are visible directly.

\subsection{Gain Attribution}
\label{sec:roles}

We ablate one component at a time on every regime (Table~\ref{tab:roles}; token-content breakdown in Supp.~\ref{app:knockout}). The decisive component is \emph{architectural} --- the per-frame within-stroke phase channel: dropping it (AU tokens only) collapses F1\textsubscript{AU} from $0.90$ to $0.79$, as the tokens name \emph{which} stroke and window but leave no anchor inside it. The per-stroke token is load-bearing in turn (the phase channel alone costs $0.06$ single-class F1\textsubscript{AU}), while the caption field it reads from the global text is redundant (within noise).

The training prior and inference guidance act on complementary axes. The off-window prior (Table~\ref{tab:roles}a) concentrates responsible-joint velocity inside the windows and helps \emph{with or without} inference guidance (both operating points in the table), so its gain is a genuine training effect, not an artefact of guidance. Inference ADG (Table~\ref{tab:roles}b) supplies recall on top, earning most where placement is hardest; its recall term is the workhorse and its off-AU term near-neutral on the final metric.

The injection \emph{pathway}, finally, outweighs raw capacity in this low-data regime: feeding both the same conditioning streams (AU tokens $+$ phase channel) and holding corpus, prior, and ADG fixed (Table~\ref{tab:roles}c), our $8.4$M gated adapter (AUG) leads the $44$M control branch (AUCN)~\cite{zhang2023controlnet} on single-class and chain placement at a fifth of the parameters, and the gap widens once inference guidance is removed ($0.87$ vs.\ $0.79$) as ADG partly compensates AUCN's weaker grounding --- grounded attention, recovering the frozen prior when no AU is supplied, fits a sparse typed condition better than a parallel branch. \label{sec:overlap}Simultaneous strokes are where per-frame text fails, and where the \emph{superposed} phase channel earns its design: the ablation pins concurrency on it (Eq.~\ref{eq:phase-field}), removing the per-frame channel costing $0.11$ on overlap ($0.77\!\to\!0.66$), nearly triple the $0.04$ from removing the AU tokens. Because each co-active stroke deposits its own content into the shared frames (count-normalized, per track) rather than overwriting, two time-overlapping strokes survive together --- which a single-token-per-frame or last-write scheme cannot provide; the discrete token set supplies the layout, the superposition their joint execution.

\begin{table}[t]
\centering\scriptsize
\setlength{\tabcolsep}{4pt}
\resizebox{\columnwidth}{!}{%
\begin{tabular}{l|ccc>{\hspace{2pt}}c}
\toprule
 & \multicolumn{3}{c}{F1\textsubscript{AU}$\uparrow$ by regime} & \\
Configuration & single-cls & chain & overlap & FID$\downarrow$ \\
\midrule
\textbf{Full ($\ell_{\mathrm{offwin}}$ $+$ AUG $+$ ADG)} & \cellcolor{oursrow}\textbf{0.899} & \cellcolor{oursrow}\textbf{0.832} & \cellcolor{oursrow}\textbf{0.770} & \cellcolor{oursrow}29.60 \\
\midrule
\multicolumn{5}{l}{\emph{(a) training streams \& prior}}\\
$-$ off-window prior ($\ell_{\mathrm{offwin}}{=}0$) & 0.868 & 0.796 & 0.700 & 30.09 \\
\quad ($\ell_{\mathrm{offwin}}{=}0$, ADG also off)  & 0.831 & 0.773 & 0.608 & --- \\
$-$ phase channel (AU tokens only)               & 0.786 & 0.739 & 0.656 & 22.75 \\
$-$ AU tokens (phase channel only)               & 0.835 & 0.809 & 0.729 & 24.23 \\
\quad $-$ caption field in AU token              & 0.902 & 0.833 & 0.746 & 27.58 \\
\midrule
\multicolumn{5}{l}{\emph{(b) inference ADG (per-term)}}\\
$-$ ADG (all terms off)                          & 0.866 & 0.807 & 0.686 & 28.63 \\
\quad $-$ recall ($\lambda_{\mathrm{pos}}{=}0$)  & 0.862 & 0.795 & 0.692 & 29.14 \\
\quad $-$ off-AU ($\lambda_{\mathrm{neg}}{=}0$)  & 0.894 & 0.835 & 0.763 & 29.51 \\
\quad $-$ DPS normalization                      & 0.900 & 0.828 & 0.770 & 30.05 \\
\midrule
\multicolumn{5}{l}{\emph{(c) conditioning pathway (base config)}}\\
AUCN (control-branch, $44$M)                     & 0.856 & 0.805 & 0.778 & 27.71 \\
\quad AUCN, ADG off                              & 0.786 & 0.745 & 0.616 & --- \\
AUG (gated, $8.4$M)                              & 0.899 & 0.832 & 0.770 & 29.60 \\
\bottomrule
\end{tabular}}
\caption{Gain attribution: one knockout per row, per regime (F1\textsubscript{AU}; AUC-E held-out detector, $\tau{=}0.5$; control metrics $3$-seed mean, FID at the canonical seed, ADG-off FID omitted). \emph{single-cls} $=$ eight-class mean, \emph{chain}/\emph{overlap} $=$ compositional sub-benches. (a) training streams $+$ off-window prior; (b) ADG per-term (recall term the workhorse); (c) AUG vs.\ the $44$M AUCN control branch. Discussion \S\ref{sec:roles}.}
\label{tab:roles}
\end{table}

\subsection{Core-Frame Position Analysis}
\label{sec:coreframe}
\begin{figure}[t]
\centering
\includegraphics[width=0.84\linewidth]{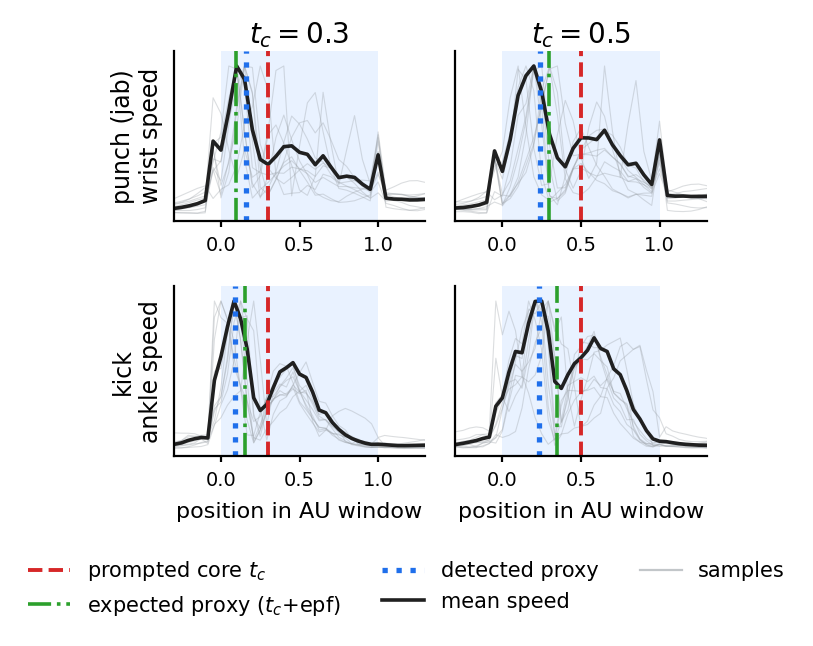}
\caption{Core-frame position as an emergent axis ($n{=}20$): rows punch (jab)/kick, columns two prompted cores ($t_c{=}0.3$/$0.6$, shaded). Each panel: responsible-joint speed (mean black, samples grey), prompted core $t_c$ (red), expected proxy $t_c{+}\delta_g$ (green), mean detected proxy (blue) --- the proxy slides with $t_c$ for kick but barely for the jab. Slopes in Table~\ref{tab:coreframe}.}
\label{fig:coreframe}
\end{figure}

We test whether the within-stroke \emph{core frame} $t_c$ is steerable. Nothing in training supervises its realized position --- $t_c$ enters only as the centre of the phase bump (\S\ref{sec:grounding}) --- so any steerability is emergent. Holding a class-typical window fixed, we prompt $t_c$ at $0.3$/$0.5$/$0.7$ of the window for the six non-postural classes ($n{=}20$) and read the realized core through the proxy--core slope $\hat s$ (\S\ref{sec:strokebench}). Five of the six track the prompt positively (Table~\ref{tab:coreframe}) --- kick most strongly ($\hat s{=}0.90$), hop ($0.77$) and throw ($0.48$) clearly, punch ($0.29$) and jump ($0.34$, apex ballistically pinned) weakly --- while \texttt{catch} does not steer ($\hat s{=}0.02$), placing its receiving instant at a near-fixed position regardless of the prompt. The velocity profiles (Fig.~\ref{fig:coreframe}) make this concrete: the detected peak slides with the prompted core for kick but barely moves for the jab. The fits are tight ($R^2{\ge}0.95$ throughout), so a moderate slope reflects loose \emph{steerability}, and \texttt{catch}'s flat-but-tight fit a consistent \emph{lack} of it, not random placement.

\begin{table}[t]
\centering\footnotesize
\begin{tabular}{lcc}
\toprule
 & slope $\hat s$ $\uparrow$ & $R^2$\\
\midrule
punch (jab) & 0.29 & 0.95\\
kick        & \textbf{0.90} & 0.99\\
jump        & 0.34 & 0.99\\
hop         & 0.77 & 0.95\\
throw       & 0.48 & 1.00\\
catch       & 0.02 & 0.96\\
\bottomrule
\end{tabular}
\caption{Core-frame controllability: proxy--core slope $\hat s$ ($\hat s{=}1$ = one-for-one) with fit $R^2$, $n{=}20$/position. Five of six non-postural classes steer; only \texttt{catch} does not. Breakdown Supp.~\ref{app:coreframe}.}
\label{tab:coreframe}
\end{table}

%% file: sec/5_discussion.tex
\section{Discussion}
\label{sec:discussion}

Per-stroke control splits cleanly into a training-side prior that carries most placement and leak suppression and an inference-side detector gradient that finishes it --- both on a \emph{frozen} backbone denoised as a single sequence, with no per-interval stitching and no retrained weights, so the gains come on top of the backbone's own motion quality. Two directions remain open: steering the realized core frame beyond the corpus's timing variance, and composing strokes that \emph{share} a track, which a per-frame detector cannot separate from fast successive chunks --- a future entity-aware detector.

%% file: sec/6_conclusion.tex
\section{Conclusion}
\label{sec:conclusion}

We promote the individual \emph{stroke} from an implicit consequence of caption-conditioned generation to an explicit, addressable condition: the Action Unit, grounded on a frozen text-to-motion backbone through a gated adapter, with Action-Detection Guidance closing residual timing errors at inference without touching a trained weight. On StrokeBench it improves per-stroke timing over caption-only, interval- and frame-level text, and joint-trajectory baselines, and the unsupervised per-stroke core frame emerges as a further steerable axis within the corpus's timing bounds.

%% file: sec/x_appendix.tex
\section{AU Corpus}
\label{app:corpus}

Beyond the per-class summary of \S\ref{sec:corpus}, candidates enter the audit pool by an OR-filter: caption-keyword match, or a sustained per-class signal from the BABEL-trained segmenter~\cite{farha2019mstcn,punnakkal2021babel} (probability $\geq 0.7$ for $\geq 5$ consecutive frames, peak above $0.5$). Velocity peaks on the responsible joint (wrist for arm tracks, ankle for leg tracks) become the per-stroke proposals; for two-leg actions (jump) the synchronized left- and right-ankle peaks become \emph{two coincident AUs} --- one per ankle track, sharing one window --- rather than one merged AU. The annotator's single-pass verdict set is pass / bad / boundary-edit / merge / track-override, and to avoid duplicate proposals an atom is dropped if its window touches any already-passed atom from an earlier-audited class (audit order: punch, kick, jump, hop, throw, catch). Table~\ref{tab:corpus} gives the per-class AUC-T/AUC-E counts and median active duration; the audit accepts $3{,}821$ Action Units across the eight classes, mirror-augmented to $7{,}642$ ($4{,}406$ AUC-T $+$ $3{,}236$ AUC-E). The postural squat is added on top of these six stroke classes through the same audit, split into a \texttt{down} and an \texttt{up} phase that are conditioned and scored as two separate sub-classes throughout (\S\ref{sec:strokebench}).

\section{Encoders and Routing: Hyperparameters}
\label{app:hparams}

\paragraph{AU encoder.}
In Eq.~\ref{eq:au-token}, $\phi$ is a Fourier expansion of each fractional frame index over $6$ frequency bands ($\sin/\cos$ at frequencies $2^0,\dots,2^5$), $\mathbf{w}_g$ and $\mathbf{a}_s$ are learned $32$-dim track/class embeddings, and $f_{\mathrm{AU}}$ is a two-layer MLP into the $D$-dimensional token. A per-slot existence flag zeroes any inactive slot's token, so it contributes nothing and the active-slot count equals the supplied stroke count.

\paragraph{Per-frame channel.}
For each frame in an active window the channel adds a duration-aware Gaussian phase bump centered at the core frame,
\begin{equation}
\varphi_i(t)=\exp\!\Big(-\tfrac{(t-t_c^i)^2}{2(\kappa\,\Delta_i)^2}\Big),\qquad \Delta_i=t_e^i-t_s^i,
\label{eq:phase}
\end{equation}
of width factor $\kappa{=}0.30$. A small MLP fuses $\varphi$ (expanded over the same $6$ Fourier bands) with four state flags (in-AU, pre-core, post-core, off-AU), the AU's track and class embeddings, and distance-to-core into a per-stroke content $\mathbf{p}_i(t)$, and the channel is the count-normalized superposition over co-active strokes (Eq.~\ref{eq:phase-field}). We keep the bump \emph{symmetric}: in a pilot comparison an asymmetric three-point profile (steep build $\sigma_b{\propto}t_c{-}t_s$, gentle decay $\sigma_d{\propto}t_e{-}t_c$) sharpened strokes at their class-typical core but tended to \emph{over-pin} the realized peak to that typical timing, slightly reducing core-frame steerability, so we adopt the simpler symmetric bump.

\paragraph{Conditioning routing.}
The conditioning adapter adds a gated self-attention block~\cite{li2023gligen} to each frozen MDM transformer layer: the layer's motion-frame tokens attend to the per-stroke AU tokens, the result is scaled by a learned gate, and the per-frame tokens add into the layer's hidden state. The gate is initialized at zero so the unconditional backbone is recovered exactly at training start; the backbone stays frozen and only the adapter ($8.4$M parameters) and AU encoder train. AU dropout $p_{\mathrm{drop}}{=}0.3$ provides classifier-free guidance. The heavier ControlNet-style control branch~\cite{zhang2023controlnet} we compare against in \S\ref{sec:roles} instead clones the full trunk ($44$M) as a locked copy with a zero-initialized output projection.

\section{Detector Features and Cross-Source Conversion}
\label{app:detector}

\paragraph{Motion representation.}
Both detectors consume the native HumanML3D~\cite{guo2022humanml3d} kinematic feature: per frame a $263$-vector of root angular/linear velocity and height ($4$), root-invariant local joint positions ($21{\times}3$), continuous-6D local joint rotations ($21{\times}6$), local joint velocities ($22{\times}3$), and binary foot-contact flags ($4$). This is exactly the representation the MDM backbone denoises, so the frozen detector reads the sampler's clean-motion estimate directly --- no conversion at guidance or scoring time --- after z-normalization by the dataset mean and standard deviation.

\paragraph{Training.}
Each detector is an offline temporal model over clips of up to $196$ frames: $d_{\mathrm{model}}{=}256$, $6$ encoder layers, and $1$ refinement stage (ASFormer additionally uses $8$ attention heads). The objective is a per-class sigmoid (multi-label binary cross-entropy) with a truncated temporal-smoothness regularizer (T-MSE, weight $0.15$) on the per-frame class probabilities, so co-active strokes on different tracks are supervised independently. We optimize with AdamW (learning rate $3{\times}10^{-4}$, weight decay $10^{-4}$) for $80$ epochs at batch size $16$, keeping the best-validation checkpoint on a held-out $10\%$ split of source clips, split by base clip id so a clip and its left/right mirror stay on the same side. The label set is the StrokeBench target classes: five laterally-typed actions (punch, kick, throw, catch, and hop --- the last split by support foot) each split left/right ($10$ channels), jump as a single two-leg class, and the squat down/up pair --- $13$ foreground channels plus one background class ($C{=}14$). The ADG guide is a C2F-TCN~\cite{singhania2023c2ftcn} and the StrokeBench evaluator an ASFormer~\cite{yi2021asformer} (\S\ref{sec:detector}).

\paragraph{Cross-source conversion.}
The generator trains on HumanML3D and most audited corpus clips are HumanML3D, but the detector corpus is supplemented with motions \emph{outside} HumanML3D (FrankenMotion/AMASS clips, for thin classes), which must be brought into the same feature space. AMASS stores SMPL pose parameters, not the $3$D joint positions HumanML3D's feature needs, so --- exactly as HumanML3D is itself built --- we run the SMPL body model (forward kinematics) to get joints and then apply HumanML3D's canonicalization: retarget to the canonical skeleton, drop onto the floor, translate the first-frame root to the $xz$ origin, and rotate so the first pose faces $+z$, then read off the $263$-vector. To check the pipeline is faithful we run it on HumanML3D clips, which already carry an official feature, and recover that feature at $\approx 0.98$ per-frame correlation; so a converted out-of-dataset clip lands in the same feature space the detector was trained on, not a subtly different convention that only resembles it.

\section{The StrokeBench Evaluator}
\label{app:heldout}

StrokeBench is scored by a single detector trained \emph{only} on the held-out AUC-E partition (HML test/val $+$ FrankenMotion, $711$ clips; Table~\ref{tab:corpus}), disjoint from the data the generator and the ADG guide train on, and a different architecture (ASFormer) from the C2F-TCN guide (\S\ref{sec:detector}), so guidance cannot inflate it. The evaluator is a competent recognizer on its held-out distribution (per-class frame-F1 $0.76$--$0.94$, mean $0.85$); its features and training recipe are in \S\ref{app:detector}.

\section{Training Objective}
\label{app:objective}

We detail the two training priors of Eq.~\ref{eq:objective}: the off-window prior $\ell_{\mathrm{offwin}}$ and the loft prior $\ell_{\mathrm{loft}}$ (\S\ref{sec:grounding}), both active in the reported model.

\paragraph{Off-window prior.} $\ell_{\mathrm{offwin}}$ ($\lambda_{\mathrm{offwin}}{=}0.8$, fixed) penalizes the squared velocity of each class's responsible joints inside every Non-AU gap (the complement of the union of that class's AU windows); suppressing this off-window motion both lowers leak and concentrates motion in the windows. Formally $\ell_{\mathrm{offwin}}=\operatorname{mean}_{t\in\mathcal{G}}\lVert\mathbf{v}_g(t)\rVert^2$ over the Non-AU gap frames $\mathcal{G}$, with $\mathbf{v}_g$ the responsible-joint velocity; the gap is the hard binary complement of the windows (the only boundary softening is the one-frame exclusion intrinsic to the finite-difference velocity). The responsible joints are the per-track wrists/ankles of Table~\ref{tab:corpus}'s classes; because a two-leg jump is stored as two coincident AUs (one per ankle track, Supp.~\ref{app:corpus}), the prior acts on \emph{both} ankle tracks and suppresses both legs outside the shared jump window. The gap and the responsible joints are defined \emph{per class}, so a mixed-class sequence (e.g.\ a punch followed by a kick) is handled correctly: the punch prior constrains only the wrist over the punch's gaps and the kick prior only the ankle over the kick's gaps, and neither penalizes the other class's stroke.

\paragraph{Loft prior.} $\ell_{\mathrm{loft}}$ ($\lambda_{\mathrm{loft}}{=}5$, fixed) supplies the one cue the velocity proxies underconstrain --- airborne \emph{height} --- and is active only for the two airborne classes (jump, hop), contributing zero for all others. For each jump/hop AU $i$ it requires the root (pelvis) height at the AU's core frame, $h(t_c^i)$, to clear the clip's off-window baseline height $\bar h_{\mathcal{G}}$ by a hinged margin $m{=}0.10$\,m:
\begin{equation}
\ell_{\mathrm{loft}}=\operatorname{mean}_{i\in\mathrm{jump,hop}} \max\!\bigl(0,\; m-(h(t_c^i)-\bar h_{\mathcal{G}})\bigr),
\end{equation}
where $\bar h_{\mathcal{G}}$ is the mean root height over that class's Non-AU gap frames (the standing baseline). The hinge makes the term one-sided --- it penalizes only \emph{insufficient} lift and does not cap a higher jump --- so it raises low or flat jumps/hops to a visibly airborne apex at the prompted instant without distorting horizontal trajectory or the velocity-driven classes. It is applied on the $\mathbf{x}_0$ estimate using the root-height channel of the HumanML3D feature (Supp.~\ref{app:detector}).

\section{Action-Detection Guidance: Terms and Stabilization}
\label{app:adg}
ADG ascends $\mathcal{J}=\lambda_{\mathrm{pos}}\mathcal{J}_{\mathrm{pos}}-\lambda_{\mathrm{neg}}\mathcal{J}_{\mathrm{neg}}$ with the detector read on the clean-motion estimate $\hat{\mathbf{x}}_0$. The recall term is a bounded margin on each prompted window's mean-log prompted-class probability, $\mathcal{J}_{\mathrm{pos}}=\sum_i-\mathrm{softplus}(\tau-\bar\ell_i)$ with $\bar\ell_i=\operatorname{mean}_{t\in W_i}\log p_{s_i}(t)$ and $\tau{=}\log 0.5$, so it saturates once a window reads as filled; the off-AU term $\mathcal{J}_{\mathrm{neg}}=\sum_{\mathrm{gaps}}\operatorname{mean}_{t\in\mathrm{gap}}(1-p_{\mathrm{bg}}(t))$ penalizes any detected action inside a gap; a count term we tried added nothing measurable and is omitted. Read on $\hat{\mathbf{x}}_0$, the gradient diverges at high noise; we normalize each term to unit norm before combining (the standard DPS device~\cite{chung2023dps}) and scale by a saturating-SNR weight, so the operating point transfers across checkpoints without retuning.

\section{Guidance Strength}
\label{app:strength}

\begin{figure}[t]
\centering
\includegraphics[width=\linewidth]{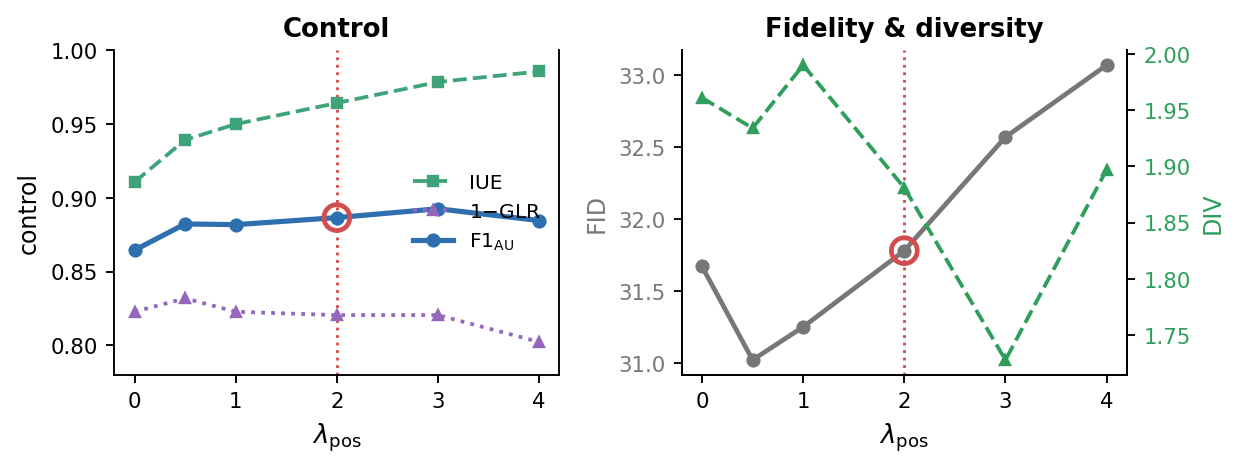}
\caption{ADG control frontier on punch ($\lambda_{\mathrm{pos}}\in\{0,0.5,1,2,3,4\}$, $\lambda_{\mathrm{neg}}{=}1.5$ fixed; $\lambda_{\mathrm{pos}}{=}0$ = recall term off), read on the held-out AUC-E detector with AUC-E FID/DIV (consistent with Table~\ref{tab:main}). The recall-term weight $\lambda_{\mathrm{pos}}$ is the only knob ($x$-axis on both panels). \textbf{Left}: control --- raw recall IUE rises monotonically ($0.91\!\to\!0.99$) while $1{-}$GLR drifts slightly down (leak grows), so their harmonic mean F1\textsubscript{AU} flattens by $\lambda_{\mathrm{pos}}{\approx}2$ ($0.887$). \textbf{Right}: FID (gray, lower is better) holds near its floor up to the knee (${\sim}31$--$32$) then climbs ($\to33.1$), while DIV (green) stays roughly flat. Control is thus bought at a modest, predictable fidelity cost that is small exactly where placement peaks, giving a clear knee; we operate at $\lambda_{\mathrm{pos}}{=}2$ (red). This sweep is punch-only at a single seed, so its absolute FID runs above Table~\ref{tab:main}'s eight-class three-seed mean; the trade-off \emph{shape} is the point. \S\ref{app:strength} reads the curve.}
\label{fig:frontier}
\end{figure}

A classifier-guided sampler usually trades fidelity for control, and the guidance strength must then be tuned per model and prompt. ADG does not erase this trade-off, but the per-sample unit-normalization (\S\ref{sec:adg}) bounds the per-step displacement at any strength, so the trade-off is gentle and monotone rather than a sharp fidelity collapse. Fig.~\ref{fig:frontier} traces it on punch, read on the held-out AUC-E detector: raw in-window recall (IUE) rises monotonically with the recall weight $\lambda_{\mathrm{pos}}$ ($0.91\!\to\!0.99$), but the placement quality F1\textsubscript{AU} --- which also charges the leak this induces --- climbs from $0.864$ and its marginal gain flattens by $\lambda_{\mathrm{pos}}{\approx}2$ ($0.887$, with no further gain past there). FID holds near its floor up to $\lambda_{\mathrm{pos}}{\approx}2$ (${\sim}31$--$32$) then climbs ($\to33.1$ at $\lambda_{\mathrm{pos}}{=}4$), and diversity stays roughly flat --- a clear knee where placement has nearly peaked but the fidelity cost is still small. We thus fix one conservative point at that knee, $(\lambda_{\mathrm{pos}},\lambda_{\mathrm{neg}}){=}(2,1.5)$, for every prompt in a single deterministic pass; being fractions of the score norm, the weights transfer across checkpoints without retuning.

\section{StrokeBench: Sub-benches and Axes}
\label{app:strokebench}

StrokeBench provides one single-class sub-bench per corpus action plus two compositional sub-benches. Single-class prompt counts: punch (13), kick (11), jump (6), hop (6), throw (6), catch (6), squat (6, both down/up phases). The punch set varies timing, count, and side. The compositional sub-benches are \emph{chain} (16 cross-class sequential prompts) and \emph{overlap} (9 concurrent cross-track prompts); Table~\ref{tab:strokebench_prompts} lists every prompt.

\paragraph{Prompt construction.} Each prompt is a \emph{programmatically constructed} AU layout, not a clip drawn from data, so none corresponds to a training motion. For a target count we place that many windows across the $196$-frame clip (cadence variants tighten or widen windows and gaps); within each window the prompted core is the class-typical fraction $t_c^i = t_s^i + \rho_c\,(t_e^i - t_s^i)$, with $\rho_c$ the median core-to-window fraction of class $c$ on the audit corpus ($\rho_{\mathrm{kick}}{\approx}0.44$), and the per-stroke pelvis target drawn with a fixed seed from class $c$'s empirical pelvis-at-core distribution. \emph{Chain} composes cross-class sequences (e.g.\ two kicks then a punch); \emph{overlap} places two strokes on \emph{different} tracks whose windows intersect (e.g.\ a punch while kicking). Each axis targets a specific observation (Table~\ref{tab:axes}).

\begin{table}[h]
\centering\footnotesize
\setlength{\tabcolsep}{4pt}
\begin{tabular}{@{}lp{6.0cm}@{}}
\toprule
Axis & Intended observation \\
\midrule
IUE / IUR & exactly-one / $\geq 1$ target chunk per window --- is the prompted count placed in-window \\
F1@.25 & segment-level match --- are strokes distinct and in the prompted order \\
GLR & fraction of Non-AU gaps with a leaked stroke --- off-window suppression \\
FID / DIV & distribution fidelity / diversity \\
\bottomrule
\end{tabular}
\caption{StrokeBench axes and the observation each is designed to surface. Single-class prompts use well-separated strokes; the overlap sub-bench places concurrent strokes on different tracks. The complete per-prompt list is given in Table~\ref{tab:strokebench_prompts}.}
\label{tab:axes}
\end{table}

\begin{figure}[t]
\centering
\includegraphics[width=\columnwidth]{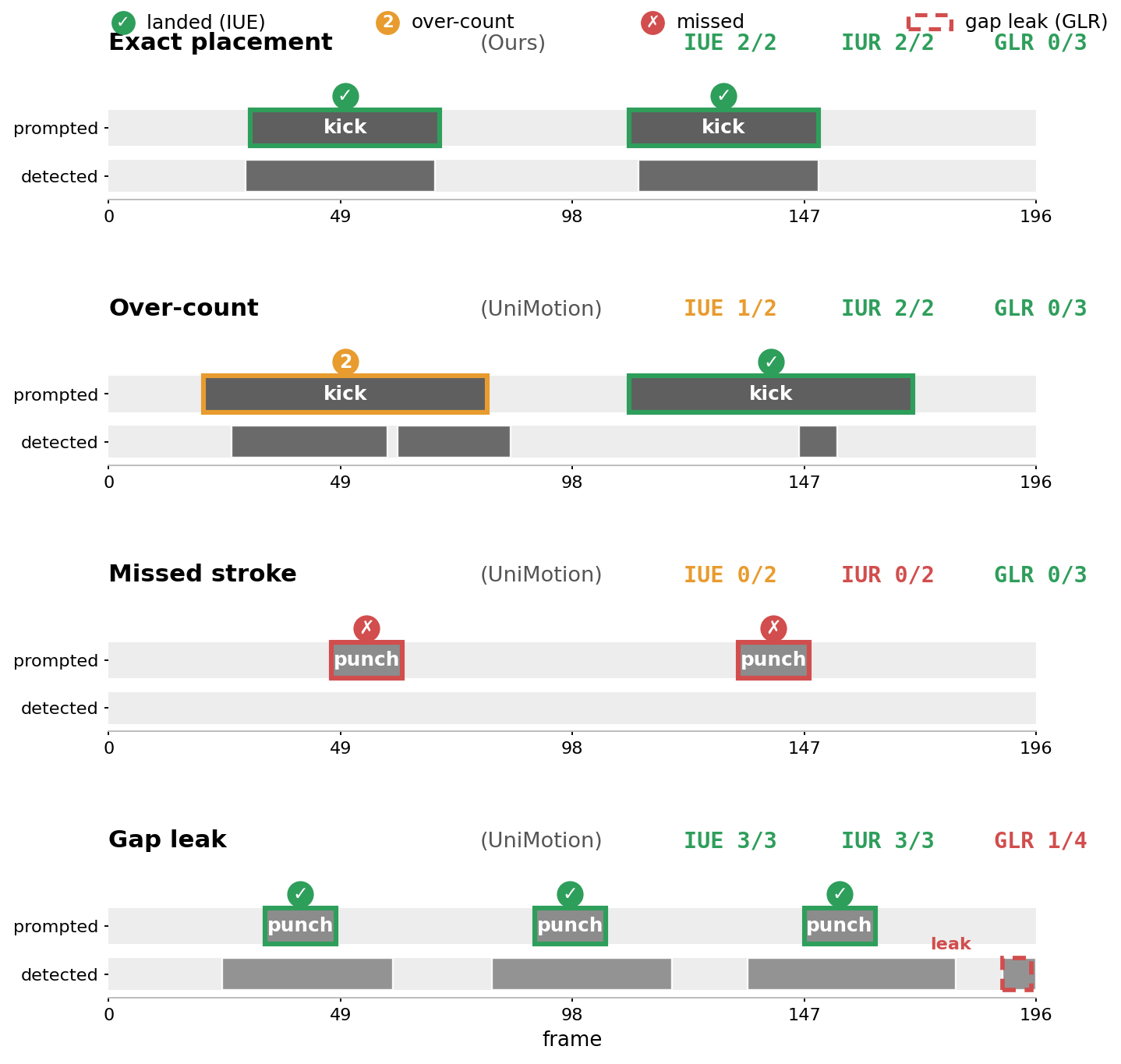}
\caption{\textbf{The held-out metric in action} on four real generated samples (one per case). We run the AUC-E evaluator ($\tau{=}0.5$) on the motion and overlay its \emph{detected chunks} (lower lane) on the \emph{prompted AU windows} (upper lane, colored by class); the badge above each window is its per-window verdict --- \textcolor{green!55!black}{$\bullet$}\,landed (exactly one overlapping chunk, counts for IUE), \textcolor{orange!85!black}{$\bullet$}\,over-count ($\geq2$ chunks), \textcolor{red!75!black}{$\bullet$}\,missed (none) --- and a dashed red box flags a chunk leaking into a Non-AU gap. \textbf{(a)} Ours places exactly one stroke per window (IUE $2/2$, no leak). \textbf{(b)} UniMotion emits an extra stroke in one window (IUE $1/2$, IUR still $2/2$ --- the count failure is what IUE catches and IUR misses). \textbf{(c)} UniMotion produces no detectable stroke (IUE/IUR $0/2$). \textbf{(d)} UniMotion hits all three windows but leaks one spurious stroke into a gap (GLR $1/4$ over the $N{+}1$ gaps). The scores track exactly the placement quality a reader sees in the timeline --- the metric is grounded in the detector's reading of the motion, not a surrogate.}
\label{fig:metric_demo}
\end{figure}

\paragraph{Unit-level axes (detector-driven).}
Both are read off the held-out detector (\S\ref{sec:detector}). Let $\mathcal{R}_c(\hat{\mathbf{x}})$ be its maximal contiguous chunks of class $c$ and $W_i=[t_s^i,t_e^i]$ for AU $i$. \textbf{IUE} (in-AU exactness, $\uparrow$) is the fraction of prompted windows with \emph{exactly one} target-class chunk overlapping; \textbf{GLR} (gap leak rate, $\downarrow$) is the fraction of a class's $N{+}1$ Non-AU gaps containing an \emph{extra} stroke of the class --- a chunk overlapping the gap by $\ge$ the minimum chunk length ($3$ frames) but no window of that class --- so a real stroke bleeding a frame past its boundary is not charged (sweeping the floor over $1$--$7$ frames moves GLR $\le0.04$ and does not reorder methods). The chunk threshold $\tau$ on the detector's per-frame class probability is calibrated \emph{once} on the held-out val split as the value maximizing mean per-class $\mathrm{HM}(\mathrm{IUE},1{-}\mathrm{GLR})$; the held-out AUC-E evaluator operates at $\tau{=}0.5$, and the guidance detector's $\tau$ is fixed independently by the same rule. The segmental \textbf{F1@.25} matches detected to prompted intervals at temporal IoU $\ge0.25$, F1 over segments~\cite{lea2017tcn}. Figure~\ref{fig:metric_demo} shows the three unit metrics computed on real evaluator outputs, one sample per failure mode. As an external check, across Table~\ref{tab:main}'s ten methods the unit F1\textsubscript{AU} rank-correlates strongly with the established segmental F1@.25 (Spearman $\rho{=}0.92$): the bespoke axes agree with a standard segmentation metric, while additionally separating the count (IUE) and off-window-leak (GLR) failure modes that a single IoU match folds together.

\section{Core-Frame Analysis}
\label{app:coreframe}

Here we add the two things the body (\S\ref{sec:coreframe}) omits: that the velocity-peak \emph{proxy} for the realized core is valid, and the operating point it is read at. Every core-frame number uses the same generator and guidance as Table~\ref{tab:main} with \emph{no} core-frame term in the guidance, so a realized core that tracks the prompt is emergent rather than steered; and the slope $\hat s$ is offset-free --- a class-typical proxy offset $\delta_g$ shifts the intercept but not the slope --- so a moderate $\hat s$ is a genuine steerability limit, not a measurement artifact.

\paragraph{Core-frame proxy validity.}
The proxy is the responsible-joint velocity peak (punch/kick/throw/catch) or the airborne-ankle height apex (jump/hop). On the audit corpus it \emph{precedes} the annotated core --- signed offsets $\delta_g=\mathbb{E}[t_{\mathrm{proxy}}{-}t_c]$ of $-4.1$ (punch), $-3.6$ (kick), $-3.2$ (catch), $-0.9$ (throw), $-0.5$ (hop), $-0.2$ (jump) frames, the responsible joint reaching peak speed just before the load-bearing instant --- and it lands within $\pm 5$ frames of that core for $71\%$ of punch, $75\%$ of kick, $98\%$ of throw, $99\%$ of jump and hop, and $68\%$ of catch. The postural squat has no velocity-peak instant of this kind, so the proxy-slope analysis is scoped to the impact and locomotion classes.

\section{Component Ablation Details}
\label{app:knockout}

This section details the from-scratch \emph{token-content} ablation behind \S\ref{sec:roles} (Table~\ref{tab:au-token}); the headline component knockouts are already in Table~\ref{tab:roles}.

\paragraph{Temporal routing of the token attention.}
The headline model leaves the token self-attention \emph{unmasked} --- each token carries its own window through its encoded boundaries (\S\ref{sec:grounding}). One can instead route the attention with an explicit temporal signal --- a \emph{hard} window mask ($t\in[t_s^i,t_e^i]$) or a \emph{soft} core-centered Gaussian bias. In our probes the two were level on \emph{chain} (F1\textsubscript{AU} $\approx0.86$) and the hard window edged the Gaussian bias on \emph{overlap} ($0.84$ vs.\ $0.82$), where disjoint windows help separate concurrent strokes, but neither improved on the simpler unmasked default, so we keep it and leave a controlled retrain to future work.

\paragraph{AU-token content.}
Given that the per-frame channel does the placing (Table~\ref{tab:roles}a), what must the token carry? Retraining the encoder from scratch with the time code changed or the token stripped to its minimal fields, phase channel fixed (Table~\ref{tab:au-token}), the token proves largely \emph{structural}: a frequency time code helps modestly (raw $-0.022$, Gaussian-RBF $-0.010$ vs.\ the $6$-band Fourier), and stripped all the way to timing $+$ count --- no track, no class --- the token still reaches F1\textsubscript{AU} $0.887$, above every baseline in Table~\ref{tab:main}. The class signal is redundant on these \emph{single-class} sub-benches, where the backbone already reads the prompted class from the global caption; it earns its place elsewhere, as the per-stroke action embedding $\mathbf{a}_s$ is what disambiguates mixed \emph{same-track} sequences (throw then catch, kick then jump, which track alone cannot separate), so we retain it. FID is flat throughout ($29$--$32$): the phase channel, not token content, governs naturalness.

\begin{table}[t]
\centering\footnotesize
\caption{AU-token content ablation (a focused content study on the six base classes; held-out AUC-E ASFormer, $\tau{=}0.5$, $n{=}10$; from-scratch retrain per row on the GAF base, single seed; phase channel and all else fixed). $\phi$: Fourier time fractions, $\mathbf{w}_g$: track, $\mathbf{a}_s$: action class, $\nu$: instance/count, $f_{\mathbf{c}}$: caption read-out. FID is the AUC-E held-out class-matched reference.}
\label{tab:au-token}
\resizebox{\columnwidth}{!}{%
\begin{tabular}{lcccc}
\toprule
AU-token content & F1\textsubscript{AU}$\uparrow$ & IUE$\uparrow$ & GLR$\downarrow$ & FID$\downarrow$\\
\midrule
Full $[\phi\,\|\,\mathbf{w}_g\,\|\,\mathbf{a}_s\,\|\,\nu\,\|\,f_{\mathbf{c}}]$ & 0.899 & 0.893 & 0.095 & 30.5\\
\midrule
\multicolumn{5}{l}{\emph{time encoding $\phi$}}\\
\quad raw fraction (no Fourier) & 0.877 & 0.867 & 0.113 & 31.8\\
\quad Gaussian-RBF (no Fourier) & 0.889 & 0.895 & 0.118 & 31.4\\
\midrule
\quad timing $+$ count only (no track/class) & 0.887 & 0.893 & 0.119 & 30.9\\
\bottomrule
\end{tabular}}
\end{table}

\section{Per-Class Scores}
\label{app:eight}

Table~\ref{tab:perclass} gives our model's per-class breakdown and Table~\ref{tab:perclass-baselines} four baselines per class; the \emph{mean} row in each is the prompt-weighted eight-class number (matching Table~\ref{tab:main}), so it need not equal the per-class average. Our single eight-class adapter places every class: the strikes near-saturate (kick $0.988$, throw $0.922$, punch $0.893$) along with squat ($0.952$), jump and catch sit a notch lower ($0.875$/$0.841$), and hop is the weakest ($0.595$) --- a sparse class that is hard for every interface. The baselines are weaker and \emph{uneven}: STMC-standstill is near-perfect on kick/throw and strong on jump but collapses on hop and squat, so its $0.76$ mean is lifted by the punch/kick-heavy weighting; UniMotion is more uniform, strong on squat and catch but below our placement on the remaining classes; Kimodo, through its full keypoint$+$text interface, is strongest on the ballistic classes (punch/kick/throw/jump, all $\geq0.91$) but collapses on the subtle hop and catch ($0.17$/$0.12$), so its $0.84$ mean rides the same punch/kick-heavy weighting. The cross-dataset FrankenMotion is the weakest overall ($0.60$) but the most \emph{uniform}: trained on AMASS body-part atoms, it alone among the baselines places hop ($0.65$) where the HumanML3D interfaces score zero, yet cannot produce a catch ($0.00$).

\begin{table}[t]
\centering
\footnotesize
\caption{Per-class scores of our model (AUG $+$ ADG, eight-class checkpoint, $N{=}10$, three seeds; held-out multi-label ASFormer at $\tau{=}0.5$; FID/DIV per class). F1\textsubscript{AU}$=$HM(IUE,$1{-}$GLR) leads, F1@.25 a trailing segmental cross-check; the \emph{mean} row is the prompt-weighted eight-class number (read with \S\ref{app:eight}).}
\label{tab:perclass}
\resizebox{\columnwidth}{!}{%
\begin{tabular}{lcccccc>{\hspace{4pt}}c}
\toprule
Class & F1\textsubscript{AU}$\uparrow$ & IUE$\uparrow$ & IUR$\uparrow$ & GLR$\downarrow$ & FID$\downarrow$ & DIV$\,\to$ & F1@.25$\uparrow$\\
\midrule
punch & 0.893 & 0.927 & 0.929 & 0.139 & 25.15 & 2.52 & 0.710\\
kick  & 0.988 & 0.984 & 0.986 & 0.008 & 27.68 & 1.59 & 0.984\\
throw & 0.922 & 0.983 & 0.983 & 0.131 & 29.40 & 1.64 & 0.777\\
jump  & 0.875 & 0.876 & 0.939 & 0.125 & 21.21 & 1.09 & 0.843\\
hop   & 0.595 & 0.433 & 0.443 & 0.051 & 48.72 & 1.76 & 0.522\\
catch & 0.841 & 0.829 & 0.833 & 0.147 & 21.96 & 2.87 & 0.719\\
squat & 0.952 & 0.942 & 0.942 & 0.037 & 20.47 & 3.04 & 0.932\\
\midrule
mean (prompt-wt.) & \cellcolor{oursrow}0.899 & 0.886 & 0.895 & 0.087 & 27.80 & 2.07 & 0.808\\
\bottomrule
\end{tabular}}
\end{table}

\begin{table*}[t]
\centering\scriptsize
\setlength{\tabcolsep}{4pt}
\caption{Per-class F1\textsubscript{AU}, IUE and GLR for four baselines (held-out AUC-E detector, $\tau{=}0.5$; \emph{mean} row prompt-weighted, matching Table~\ref{tab:main}): the three strongest HumanML3D-trained interfaces (STMC-standstill, UniMotion, Kimodo, $n{=}10$) and the cross-dataset FrankenMotion reference ($^\dagger$ $1$ deterministic motion/prompt). All are uneven across classes --- read with the text (\S\ref{app:eight}).}
\label{tab:perclass-baselines}
\begin{tabular}{lccc@{\hspace{6pt}}ccc@{\hspace{6pt}}ccc@{\hspace{6pt}}ccc}
\toprule
 & \multicolumn{3}{c}{STMC (standstill)} & \multicolumn{3}{c}{UniMotion} & \multicolumn{3}{c}{Kimodo (full)} & \multicolumn{3}{c}{FrankenMotion$^{\dagger}$}\\
\cmidrule(lr){2-4}\cmidrule(lr){5-7}\cmidrule(lr){8-10}\cmidrule(lr){11-13}
Class & F1\textsubscript{AU} & IUE & GLR & F1\textsubscript{AU} & IUE & GLR & F1\textsubscript{AU} & IUE & GLR & F1\textsubscript{AU} & IUE & GLR\\
\midrule
punch & 0.828 & 0.768 & 0.102 & 0.725 & 0.593 & 0.066 & 0.926 & 0.986 & 0.127 & 0.609 & 0.464 & 0.114\\
kick  & 1.000 & 1.000 & 0.000 & 0.840 & 0.724 & 0.000 & 0.908 & 0.838 & 0.009 & 0.542 & 0.429 & 0.265\\
throw & 0.889 & 1.000 & 0.200 & 0.899 & 0.930 & 0.129 & 0.930 & 0.950 & 0.088 & 0.803 & 0.700 & 0.059\\
jump  & 0.947 & 0.900 & 0.000 & 0.860 & 0.755 & 0.000 & 1.000 & 1.000 & 0.000 & 0.739 & 0.636 & 0.118\\
hop   & 0.000 & 0.000 & 0.000 & 0.000 & 0.000 & 0.000 & 0.165 & 0.090 & 0.018 & 0.653 & 0.500 & 0.059\\
catch & 0.605 & 0.438 & 0.020 & 0.793 & 0.838 & 0.247 & 0.118 & 0.062 & 0.000 & 0.000 & 0.000 & 0.133\\
squat & 0.137 & 0.075 & 0.215 & 0.992 & 1.000 & 0.015 & 0.809 & 0.725 & 0.085 & 0.621 & 0.562 & 0.308\\
\midrule
mean (prompt-wt.) & 0.758 & 0.645 & 0.081 & 0.800 & 0.693 & 0.054 & 0.839 & 0.757 & 0.058 & 0.602 & 0.471 & 0.165\\
\bottomrule
\end{tabular}
\end{table*}

\section{Conditioning Generality on a Frozen UniMotion}
\label{app:unimotion}\label{sec:unimotion}

AU grounding is not tied to MDM. The adapter injects zero-gated attention into a backbone's frozen transformer blocks (\S\ref{sec:grounding}), so it ports to any frozen \emph{transformer} text-to-motion denoiser without retraining its weights. We attach the \emph{same} adapter ($13.2$M trainable) unchanged onto a frozen UniMotion~\cite{li2025unimotion} --- a stronger $1000$-step transformer-diffusion model that, unlike MDM, already exposes a \emph{native} per-frame text interface --- training only the adapter on AUC-T while UniMotion stays frozen. The zero-initialized gate makes the AU-off pass reproduce the backbone exactly, so every change is attributable to the adapter alone.

\begin{table}[t]
\centering\scriptsize
\setlength{\tabcolsep}{3pt}
\caption{Conditioning generality, prompt-weighted (micro) mean over the six \emph{stroke} classes (held-out AUC-E detector, $\tau{=}0.5$; class-matched AUC-E FID). The \emph{same} adapter on \emph{two} frozen HumanML3D backbones improves per-stroke placement over each backbone's native conditioning. ``$+$adapter'' rows isolate the adapter (no ADG); the $+$ADG MDM row is Table~\ref{tab:main}'s ``Ours''. The postural \emph{squat} is the one class on which the adapter \emph{regresses} a frozen UniMotion (it shallows its native deep descent; see text), so it is excluded here and discussed separately. F1\textsubscript{AU}$=$HM(IUE,$1{-}$GLR).}
\label{tab:unimotion}
\begin{tabular}{@{}ll@{\hspace{4pt}}ccccc@{}}
\toprule
Backbone & Cond. & ADG & F1\textsubscript{AU}$\uparrow$ & IUE$\uparrow$ & GLR$\downarrow$ & FID$\downarrow$\\
\midrule
\multirow{2}{*}{UniMotion} & native text & --- & 0.759 & 0.637 & \textbf{0.061} & 23.8\\
 & $+$adapter & --- & 0.814 & 0.720 & 0.065 & \textbf{21.9}\\
\midrule
\multirow{2}{*}{MDM} & $+$adapter & --- & 0.859 & 0.860 & 0.142 & 30.1\\
 & $+$adapter & \checkmark & \textbf{0.899} & \textbf{0.893} & 0.095 & 30.5\\
\bottomrule
\end{tabular}
\end{table}

For the \emph{native} baseline we hand UniMotion our AU timeline as per-frame text (each positive-window frame ``\texttt{<side> <action>}'', gaps a rest phrase). The adapter lifts native UniMotion across the six stroke classes (F1\textsubscript{AU} $0.759\!\to\!0.814$, IUE $0.637\!\to\!0.720$) while \emph{improving} fidelity (FID $23.8\!\to\!21.9$), the gain carried by the ballistic strikes per-frame text underspecifies (kick $0.84\!\to\!0.99$, jump $0.86\!\to\!0.95$). The one regression is the postural \emph{squat}: the timing-driven adapter shallows UniMotion's native deep descent (root drop ${\sim}$half), which the held-out detector no longer reads as a squat, so we report it separately and exclude it above. The \emph{same} module thus makes both a caption-only (MDM) and a frame-level-text (UniMotion) backbone per-stroke-controllable --- AU grounding is a reusable interface, not an MDM-specific trick. \emph{Implementation:} we install the gated attention and phase channel as per-layer forward hooks on the frozen UniMotion transformer (skipping its three prefix tokens vs.\ MDM's one) and train only the $13.2$M adapter, which \emph{augments} rather than replaces UniMotion's native text. We run it without ADG, whose $50$-step operating point is diluted ${\sim}20\times$ on the $1000$-step schedule and effectively inert.

\section{Converting StrokeBench Prompts to Baseline Inputs}
\label{app:stmc}

Each baseline receives the \emph{same} AU set through its own native interface; the only glue is mapping the StrokeBench AU windows to that method's input format.
\begin{itemize}\setlength{\itemsep}{2pt}\setlength{\parskip}{0pt}
\item \textbf{FineMoGen}: native temporal composition --- one text segment per AU window carrying that window's \texttt{motion\_length}, plus a gap segment for each silent interval (no in-text time string).
\item \textbf{STMC}: a body-part timeline (\texttt{<text>\#<t\_s>\#<t\_e>\#<part>} per positive AU on track $g$, handedness-aware singular sub-caption). Its stitcher must cover $[0,T]$, so we report two ways of handling the gaps: \emph{standstill} inserts an explicit ``a person stands still'' caption per gap, and \emph{overlay} composes the per-stroke captions over an empty-caption full-body base.
\item \textbf{FrankenMotion}: per-body-part text with time windows; it responds to in-vocabulary phrasing, and count phrasings such as ``a punch''/``one punch'' are out-of-vocabulary (the gerund ``punching'' appears ${\sim}358\times$ in its training annotations vs.\ $0$ for ``a punch''), which we verified yields a weaker, less reliable stroke (${\sim}0.6\times$ the in-window arm speed of the gerund), so positives use the in-vocabulary gerunds (``punching'', etc.) and gaps use its no-constraint token. Trained on a different (AMASS body-part) dataset, so reported as a cross-dataset reference only.
\item \textbf{UniMotion}: a dense per-frame text track (we supply per-frame action-class labels from the AU windows). \textbf{DART}: a sequence of text-labeled autoregressive motion primitives. Both are HumanML3D-trained.
\item \textbf{OmniControl, Kimodo} (joint trajectory): \textbf{OmniControl} controls positions, not text timing, so we hand it \emph{oracle} per-stroke keypoints at the prompted core frames --- the responsible joint's position at the velocity peak (arm$\to$wrist, leg$\to$ankle), averaged over that class's real GT clips, one keypoint per positive AU --- written into its dense spatial-hint tensor (HumanML3D-native, 22-joint; only that (frame, joint) active). \textbf{Kimodo} natively combines text with kinematic constraints, so we exercise its \emph{full} interface: an interval-level text timeline (a subject-first sub-caption per AU window, e.g.\ ``A person throws a right punch'', with ``A person stands still'' on the gaps) conditioned \emph{jointly} with one per-limb keyframe per AU at its core frame (seeded with a canonical pose from a text-only Kimodo generation of that class, left-side AUs mirrored), through Kimodo's multi-prompt$+$constraint API with separate text/constraint guidance. It uses Kimodo's \mbox{Llama-3-8B} LLM2Vec encoder and SMPL-X 22-joint model, runs at $30$\,fps (core indices $\times1.5$, output resampled to $20$\,fps), re-normalized to our statistics. This full interface is far stronger than keyframes alone --- its eight-class IUE ($0.76$, Table~\ref{tab:main}) sits well above the keyframes-only constraint setup. Because its SMPL-X outputs are retargeted to our HumanML3D skeleton (a lossy conversion) and the heavily constrained generations drift off-distribution for some classes (catch especially), we read its FID as a loose upper bound.
\end{itemize}

\paragraph{Laterality.} The captions are side-agnostic (``three punches''), so side is carried by the AU's track $g$, and every conversion above passes it through its own side-bearing channel: the handed sub-caption (STMC, FineMoGen, Kimodo, DART), the per-frame \texttt{<side> <action>} label (UniMotion), the body part (FrankenMotion), or the responsible-joint keypoint (OmniControl). Every interface is thus given the prompted side, and the side-typed detector scores each against it; only the caption-only MDM, with no side channel, cannot receive it.

We draw $N{=}10$ samples per prompt (re-seeding the deterministic methods). Every generated motion is mapped to the HumanML3D 263-d feature space and re-normalized to our statistics before the held-out detector / FID / DIV.

\section{Baseline Inference Cost}
\label{app:cost}
Table~\ref{tab:cost} lists each baseline's generator size and sampling budget. Inference wall-clock is dominated by the number of denoiser passes (sampling steps): Ours and the MDM backbone use a $50$-step schedule, whereas UniMotion and OmniControl run the full $1000$-step DDPM --- a $20\times$ deeper sampling loop per motion --- so UniMotion's per-stroke gains in \S\ref{sec:unimotion} come at a substantially higher inference cost than our $50$-step generation. Ours adds only a $13.2$M adapter on top of a \emph{frozen} $23.0$M MDM, training no backbone weights.

\begin{table}[t]
\centering
\footnotesize
\caption{Generator parameter count and sampling budget per baseline (HumanML3D inference setting). Params are the generative model used at inference (DART $=$ denoiser $+$ motion VAE) and \emph{exclude each method's text encoder} --- CLIP for the diffusion models, and for Kimodo the frozen ${\sim}8$B Llama-3-8B LLM2Vec encoder (run per caption, not per denoiser step). $^\dagger$Ours trains only the adapter; the $23.0$M MDM backbone is frozen. $^\ddagger$Kimodo's $282$\,M is its diffusion denoiser only. \emph{Latency}: wall-clock for a single motion on one A6000 (model load excluded), each method at its evaluated operating point; the cost \emph{structure} differs by method. Ours' $1.5$\,s is one global $50$-step pass --- bare backbone ${\sim}0.2$\,s plus ADG's \emph{per-step} detector gradient (${\sim}1.3$\,s over the $50$ steps); the $1000$-step baselines are one global pass at high step count (UniMotion $22$\,s; OmniControl $74$\,s, with dense \emph{per-step} spatial guidance). Kimodo's ${\sim}11$\,s instead scales \emph{per timeline segment}: ${\sim}3.2$\,s fixed $+$ ${\sim}1.5$\,s/segment (one $100$-step DDIM pass each; $3$--$9$ segments per StrokeBench prompt $=7.7$--$17.7$\,s, a single-caption motion only ${\sim}5$\,s), of which the Llama-3-8B encode is ${\sim}5\%$ (${\sim}0.15$\,s/segment) and the rest denoising.}
\label{tab:cost}
\resizebox{\columnwidth}{!}{%
\begin{tabular}{lcccc}
\toprule
Method & Params (M) & Steps & Sampler & Latency (s)\\
\midrule
MDM~\cite{tevet2023mdm}            & 23.0 & 50   & DDPM & 0.2\\
FineMoGen~\cite{zhang2023finemogen}& 65.5 & 50   & DDIM & 1.6\\
UniMotion~\cite{li2025unimotion}   & 25.1 & 1000 & DDPM & 22\\
DART~\cite{zhao2025dartcontrol}    & 35.5 & 10   & latent AR & 2.5\\
STMC~\cite{petrovich2024stmc}  & 27.0 & 100  & DDPM & 0.8\\
FrankenMotion~\cite{li2026frankenmotion} & 14.8 & 100 & DDPM & 0.2\\
OmniControl~\cite{xie2024omnicontrol}    & 48.8 & 1000 & DDPM & 74\\
Kimodo~\cite{kimodo2026}           & $282^{\ddagger}$  & 100  & DDIM & 11\\
\midrule
\textbf{Ours}        & $23.0{+}\mathbf{13.2}^{\dagger}$ & 50 & DDPM & 1.5\\
\bottomrule
\end{tabular}}
\end{table}

\section{StrokeBench Prompt Listing}
\label{app:prompts}
The complete per-prompt StrokeBench specification (Table~\ref{tab:strokebench_prompts}), referenced from Supp.~\ref{app:strokebench}.

\onecolumn
\input{tab_strokebench_prompts}
\twocolumn

%% file: tab_strokebench_prompts.tex
\begingroup\footnotesize
\begin{longtable}{@{}p{1.3cm}p{0.80\textwidth}@{}}
\caption{Complete StrokeBench prompt list. For each prompt the first line is its Action-Unit layout --- level, stroke count ($\times$), body track(s), and per-stroke window(s) (start--end, prompted core frame in parentheses; 196-frame clips) --- followed by the prompt. Tracks: RA/LA right/left arm, RL/LL right/left leg, T trajectory/torso. The per-class blocks prompt a single action class (possibly several strokes of it; squat annotates its down/up phases); the \emph{chain} block holds sequential cross-class sequences (strokes non-overlapping in time) and the \emph{overlap} block concurrent pairs (two strokes co-active on different tracks).}\label{tab:strokebench_prompts}\\
\toprule Class & Action-Unit layout \,/\, prompt \\ \midrule
\endfirsthead
\multicolumn{2}{@{}l}{\footnotesize\itshape Table~\thetable, continued}\\[2pt] \toprule Class & Action-Unit layout \,/\, prompt \\ \midrule
\endhead
punch & \textbf{one}\quad $1\times$\quad RA\,[86--110 (98)]\newline \textit{a person throws a single punch} \\
\addlinespace[2pt]
 & \textbf{two}\quad $2\times$\quad RA\,[47--62 (54); 133--148 (140)]\newline \textit{a person throws two punches} \\
\addlinespace[2pt]
 & \textbf{three}\quad $3\times$\quad RA\,[33--48 (40); 90--105 (98); 147--162 (154)]\newline \textit{a person throws three punches in succession} \\
\addlinespace[2pt]
 & \textbf{four}\quad $4\times$\quad RA\,[26--41 (34); 69--84 (76); 112--127 (120); 155--170 (162)]\newline \textit{a person throws four quick punches} \\
\addlinespace[2pt]
 & \textbf{rapid}\quad $3\times$\quad RA\,[40--50 (45); 70--80 (75); 100--110 (105)]\newline \textit{a person rapidly punches three times} \\
\addlinespace[2pt]
 & \textbf{deliberate}\quad $2\times$\quad RA\,[30--80 (55); 110--160 (135)]\newline \textit{a person throws two slow deliberate punches} \\
\addlinespace[2pt]
 & \textbf{after}\quad $1\times$\quad RA\,[40--60 (50)]\newline \textit{a person throws one punch and then stands still} \\
\addlinespace[2pt]
 & \textbf{before}\quad $1\times$\quad RA\,[120--150 (135)]\newline \textit{a person stands still then throws one punch} \\
\addlinespace[2pt]
 & \textbf{between}\quad $2\times$\quad RA\,[30--50 (40); 130--150 (140)]\newline \textit{a person throws one punch, pauses, then throws one more} \\
\addlinespace[2pt]
 & \textbf{left}\quad $1\times$\quad LA\,[86--110 (98)]\newline \textit{a person throws a punch with their left fist} \\
\addlinespace[2pt]
 & \textbf{left\_then\_right}\quad $2\times$\quad LA\,[30--60 (45)]\quad RA\,[110--140 (125)]\newline \textit{a person throws a left jab and then a right cross} \\
\addlinespace[2pt]
 & \textbf{right\_then\_left}\quad $2\times$\quad RA\,[30--60 (45)]\quad LA\,[110--140 (125)]\newline \textit{a person throws a right jab and then a left cross} \\
\addlinespace[2pt]
 & \textbf{lrlr}\quad $4\times$\quad LA\,[20--40 (30); 100--120 (110)]\quad RA\,[60--80 (70); 140--160 (150)]\newline \textit{a person alternates left and right jabs four times} \\
\midrule
kick & \textbf{one}\quad $1\times$\quad RL\,[60--110 (82)]\newline \textit{a person kicks once and stops} \\
\addlinespace[2pt]
 & \textbf{two}\quad $2\times$\quad RL\,[30--70 (48); 110--150 (128)]\newline \textit{a person kicks twice} \\
\addlinespace[2pt]
 & \textbf{three}\quad $3\times$\quad RL\,[20--50 (33); 70--100 (83); 120--150 (133)]\newline \textit{a person kicks three times in succession} \\
\addlinespace[2pt]
 & \textbf{rapid}\quad $3\times$\quad RL\,[30--55 (41); 70--95 (81); 110--135 (121)]\newline \textit{a person quickly kicks three times} \\
\addlinespace[2pt]
 & \textbf{deliberate}\quad $2\times$\quad RL\,[20--80 (46); 110--170 (136)]\newline \textit{a person delivers two slow deliberate kicks} \\
\addlinespace[2pt]
 & \textbf{after}\quad $1\times$\quad RL\,[40--80 (58)]\newline \textit{a person kicks once and then stands still} \\
\addlinespace[2pt]
 & \textbf{before}\quad $1\times$\quad RL\,[110--150 (128)]\newline \textit{a person stands still and then kicks once} \\
\addlinespace[2pt]
 & \textbf{left}\quad $1\times$\quad LL\,[60--110 (82)]\newline \textit{a person kicks with their left leg} \\
\addlinespace[2pt]
 & \textbf{right}\quad $1\times$\quad RL\,[60--110 (82)]\newline \textit{a person kicks with their right leg} \\
\addlinespace[2pt]
 & \textbf{left\_then\_right}\quad $2\times$\quad LL\,[30--70 (48)]\quad RL\,[110--150 (128)]\newline \textit{a person kicks with their left leg, then with their right leg} \\
\addlinespace[2pt]
 & \textbf{lrlr}\quad $4\times$\quad LL\,[20--45 (31); 100--125 (111)]\quad RL\,[60--85 (71); 140--165 (151)]\newline \textit{a person alternates left and right kicks four times} \\
\midrule
jump & \textbf{one}\quad $1\times$\quad RL\,[60--105 (80)]\newline \textit{a person jumps once} \\
\addlinespace[2pt]
 & \textbf{two}\quad $2\times$\quad RL\,[40--75 (55); 110--145 (125)]\newline \textit{a person jumps twice} \\
\addlinespace[2pt]
 & \textbf{three}\quad $3\times$\quad RL\,[25--55 (38); 75--105 (88); 125--155 (138)]\newline \textit{a person jumps three times in a row} \\
\addlinespace[2pt]
 & \textbf{high}\quad $1\times$\quad RL\,[40--100 (66)]\newline \textit{a person makes a high jump} \\
\addlinespace[2pt]
 & \textbf{rapid}\quad $3\times$\quad RL\,[35--55 (44); 75--95 (84); 115--135 (124)]\newline \textit{a person makes small quick jumps three times} \\
\addlinespace[2pt]
 & \textbf{before}\quad $1\times$\quad RL\,[110--155 (130)]\newline \textit{a person stands still and then jumps once} \\
\midrule
hop & \textbf{one}\quad $1\times$\quad RL\,[60--90 (74)]\newline \textit{a person hops once on their right foot} \\
\addlinespace[2pt]
 & \textbf{two}\quad $2\times$\quad RL\,[40--65 (52); 110--135 (122)]\newline \textit{a person hops twice on the same foot} \\
\addlinespace[2pt]
 & \textbf{three}\quad $3\times$\quad RL\,[30--55 (42); 75--100 (86); 120--145 (132)]\newline \textit{a person hops three times in a row} \\
\addlinespace[2pt]
 & \textbf{left}\quad $1\times$\quad LL\,[60--90 (74)]\newline \textit{a person hops on their left foot} \\
\addlinespace[2pt]
 & \textbf{right}\quad $1\times$\quad RL\,[60--90 (74)]\newline \textit{a person hops on their right foot} \\
\addlinespace[2pt]
 & \textbf{left\_then\_right}\quad $2\times$\quad LL\,[40--65 (52)]\quad RL\,[110--135 (122)]\newline \textit{a person hops on the left foot and then on the right foot} \\
\midrule
throw & \textbf{one}\quad $1\times$\quad RA\,[50--100 (72)]\newline \textit{a person throws a ball once} \\
\addlinespace[2pt]
 & \textbf{two}\quad $2\times$\quad RA\,[30--70 (48); 110--150 (128)]\newline \textit{a person throws a ball twice} \\
\addlinespace[2pt]
 & \textbf{three}\quad $3\times$\quad RA\,[20--55 (35); 70--105 (85); 120--155 (135)]\newline \textit{a person throws three balls in a row} \\
\addlinespace[2pt]
 & \textbf{left}\quad $1\times$\quad LA\,[50--100 (72)]\newline \textit{a person throws a ball with their left arm} \\
\addlinespace[2pt]
 & \textbf{right}\quad $1\times$\quad RA\,[50--100 (72)]\newline \textit{a person throws a ball with their right arm} \\
\addlinespace[2pt]
 & \textbf{left\_then\_right}\quad $2\times$\quad LA\,[30--70 (48)]\quad RA\,[110--150 (128)]\newline \textit{a person throws a ball first with their left arm, then with their right} \\
\midrule
catch & \textbf{one}\quad $1\times$\quad RA\,[60--100 (80)]\newline \textit{a person catches a ball once} \\
\addlinespace[2pt]
 & \textbf{two}\quad $2\times$\quad RA\,[40--70 (55); 110--140 (125)]\newline \textit{a person catches a ball twice} \\
\addlinespace[2pt]
 & \textbf{left}\quad $1\times$\quad LA\,[60--100 (80)]\newline \textit{a person catches a ball with their left hand} \\
\addlinespace[2pt]
 & \textbf{right}\quad $1\times$\quad RA\,[60--100 (80)]\newline \textit{a person catches a ball with their right hand} \\
\addlinespace[2pt]
 & \textbf{left\_then\_right}\quad $2\times$\quad LA\,[40--70 (55)]\quad RA\,[110--140 (125)]\newline \textit{a person catches with the left then the right hand} \\
\addlinespace[2pt]
 & \textbf{before}\quad $1\times$\quad RA\,[110--150 (130)]\newline \textit{a person stands still and then catches a ball} \\
\midrule
squat & \textbf{one}\quad $2\times$\quad T\,[55--90 (72) squat\,$\downarrow$; 90--125 (107) squat\,$\uparrow$]\newline \textit{a person squats down and stands up once} \\
\addlinespace[2pt]
 & \textbf{two}\quad $4\times$\quad T\,[35--60 (47) squat\,$\downarrow$; 60--85 (72) squat\,$\uparrow$; 105--130 (117) squat\,$\downarrow$; 130--155 (142) squat\,$\uparrow$]\newline \textit{a person squats twice} \\
\addlinespace[2pt]
 & \textbf{three}\quad $6\times$\quad T\,[25--45 (35) squat\,$\downarrow$; 45--65 (55) squat\,$\uparrow$; 75--95 (85) squat\,$\downarrow$; 95--115 (105) squat\,$\uparrow$; 125--145 (135) squat\,$\downarrow$; 145--165 (155) squat\,$\uparrow$]\newline \textit{a person does three squats in a row} \\
\addlinespace[2pt]
 & \textbf{deep}\quad $2\times$\quad T\,[50--95 (72) squat\,$\downarrow$; 95--140 (117) squat\,$\uparrow$]\newline \textit{a person does one deep squat} \\
\addlinespace[2pt]
 & \textbf{down\_only}\quad $1\times$\quad T\,[60--105 (80) squat\,$\downarrow$]\newline \textit{a person squats down and holds the position} \\
\addlinespace[2pt]
 & \textbf{up\_only}\quad $1\times$\quad T\,[60--105 (80) squat\,$\uparrow$]\newline \textit{a person stands up from a squatting position} \\
\midrule
chain & \textbf{punch\_LR}\quad $2\times$\quad LA\,[35--49 (42) punch]\quad RA\,[100--114 (107) punch]\newline \textit{a person punches with the left hand, then punches with the right hand} \\
\addlinespace[2pt]
 & \textbf{punch\_LRL}\quad $3\times$\quad LA\,[25--39 (32) punch; 135--149 (142) punch]\quad RA\,[80--94 (87) punch]\newline \textit{a person punches with the left hand, then punches with the right hand, then punches with the left hand} \\
\addlinespace[2pt]
 & \textbf{punch\_RLRL}\quad $4\times$\quad RA\,[20--32 (26) punch; 104--116 (110) punch]\quad LA\,[62--74 (68) punch; 146--158 (152) punch]\newline \textit{a person punches with the right hand, then punches with the left hand, then punches with the right hand, then punches with the left hand} \\
\addlinespace[2pt]
 & \textbf{kick\_LR}\quad $2\times$\quad LL\,[35--55 (45) kick]\quad RL\,[105--125 (115) kick]\newline \textit{a person kicks with the left leg, then kicks with the right leg} \\
\addlinespace[2pt]
 & \textbf{kick\_RLR}\quad $3\times$\quad RL\,[25--45 (35) kick; 135--155 (145) kick]\quad LL\,[80--100 (90) kick]\newline \textit{a person kicks with the right leg, then kicks with the left leg, then kicks with the right leg} \\
\addlinespace[2pt]
 & \textbf{kick\_jump}\quad $2\times$\quad RL\,[30--50 (40) kick; 100--118 (109) jump]\newline \textit{a person kicks with the right leg, then jumps} \\
\addlinespace[2pt]
 & \textbf{jump\_kick}\quad $2\times$\quad RL\,[30--48 (39) jump; 100--120 (110) kick]\newline \textit{a person jumps, then kicks with the right leg} \\
\addlinespace[2pt]
 & \textbf{kick\_jump\_kick}\quad $3\times$\quad RL\,[20--40 (30) kick; 78--96 (87) jump; 135--155 (145) kick]\newline \textit{a person kicks with the right leg, then jumps, then kicks with the right leg} \\
\addlinespace[2pt]
 & \textbf{throw\_catch}\quad $2\times$\quad RA\,[35--53 (44) throw; 105--123 (114) catch]\newline \textit{a person throws a ball with the right hand, then catches a ball with the right hand} \\
\addlinespace[2pt]
 & \textbf{catch\_throw}\quad $2\times$\quad RA\,[35--53 (44) catch; 105--123 (114) throw]\newline \textit{a person catches a ball with the right hand, then throws a ball with the right hand} \\
\addlinespace[2pt]
 & \textbf{punch\_throw}\quad $2\times$\quad RA\,[35--49 (42) punch; 100--118 (109) throw]\newline \textit{a person punches with the right hand, then throws a ball with the right hand} \\
\addlinespace[2pt]
 & \textbf{punch\_kick}\quad $2\times$\quad RA\,[30--44 (37) punch]\quad RL\,[100--120 (110) kick]\newline \textit{a person punches with the right hand, then kicks with the right leg} \\
\addlinespace[2pt]
 & \textbf{kick\_punch}\quad $2\times$\quad RL\,[30--50 (40) kick]\quad RA\,[105--119 (112) punch]\newline \textit{a person kicks with the right leg, then punches with the right hand} \\
\addlinespace[2pt]
 & \textbf{kjp}\quad $3\times$\quad RL\,[20--40 (30) kick; 72--90 (81) jump]\quad RA\,[135--149 (142) punch]\newline \textit{a person kicks with the right leg, then jumps, then punches with the right hand} \\
\addlinespace[2pt]
 & \textbf{p3\_kick}\quad $4\times$\quad RA\,[20--32 (26) punch; 50--62 (56) punch; 80--92 (86) punch]\quad RL\,[130--150 (140) kick]\newline \textit{a person punches with the right hand, then punches with the right hand, then punches with the right hand, then kicks with the right leg} \\
\addlinespace[2pt]
 & \textbf{lpunch\_rpunch\_jump}\quad $3\times$\quad LA\,[25--39 (32) punch]\quad RA\,[75--89 (82) punch]\quad RL\,[130--148 (139) jump]\newline \textit{a person punches with the left hand, then punches with the right hand, then jumps} \\
\midrule
overlap & \textbf{punchRA\_kickRL\_sim}\quad $2\times$\quad RA\,[70--90 (80) punch]\quad RL\,[70--94 (82) kick]\newline \textit{a person punches with the right hand and kicks with the right leg, at the same time} \\
\addlinespace[2pt]
 & \textbf{Lpunch\_Rkick\_sim}\quad $2\times$\quad RL\,[68--94 (81) kick]\quad LA\,[70--86 (78) punch]\newline \textit{a person kicks with the right leg and punches with the left hand, at the same time} \\
\addlinespace[2pt]
 & \textbf{punch\_jump\_sim}\quad $2\times$\quad RL\,[58--82 (70) jump]\quad RA\,[60--78 (69) punch]\newline \textit{a person jumps and punches with the right hand, at the same time} \\
\addlinespace[2pt]
 & \textbf{throwRA\_kickRL\_sim}\quad $2\times$\quad RA\,[60--80 (70) throw]\quad RL\,[62--88 (75) kick]\newline \textit{a person throws a ball with the right hand and kicks with the right leg, at the same time} \\
\addlinespace[2pt]
 & \textbf{twohand\_punch\_sim}\quad $2\times$\quad LA\,[70--88 (79) punch]\quad RA\,[70--88 (79) punch]\newline \textit{a person punches with the left hand and punches with the right hand, at the same time} \\
\addlinespace[2pt]
 & \textbf{punch\_then\_simkick}\quad $3\times$\quad RA\,[30--46 (38) punch; 95--113 (104) punch]\quad RL\,[95--119 (107) kick]\newline \textit{a person punches with the right hand, then punches with the right hand and kicks with the right leg, at the same time} \\
\addlinespace[2pt]
 & \textbf{sim\_then\_seq}\quad $3\times$\quad RA\,[40--60 (50) punch; 120--136 (128) punch]\quad RL\,[42--66 (54) kick]\newline \textit{a person punches with the right hand and kicks with the right leg, at the same time, then punches with the right hand} \\
\addlinespace[2pt]
 & \textbf{punch\_kick\_ovl}\quad $2\times$\quad RA\,[50--75 (62) punch]\quad RL\,[68--95 (81) kick]\newline \textit{a person punches with the right hand and kicks with the right leg, at the same time} \\
\addlinespace[2pt]
 & \textbf{kick\_punch\_ovl}\quad $2\times$\quad RL\,[50--80 (65) kick]\quad RA\,[72--96 (84) punch]\newline \textit{a person kicks with the right leg and punches with the right hand, at the same time} \\
\bottomrule
\end{longtable}\endgroup